\documentclass[preprint,12pt]{elsarticle}

\usepackage{amsmath,amssymb,amsfonts}
\usepackage{algorithmic}
\usepackage{graphicx}
\usepackage{textcomp}
\usepackage{xcolor}

\usepackage[final]{hyperref}
\usepackage{comment}
\usepackage{soul}
\usepackage{mathtools}
\usepackage{paralist}
\usepackage{booktabs}
\usepackage{caption}
\usepackage{subcaption}
\usepackage{verbatim}
\usepackage{siunitx}
\usepackage[flushleft]{threeparttable}
\usepackage{natbib}

\newtheorem{definition}{Definition}

\newcommand\btext[1]{{\color{black}{#1}}}
\newcommand\bbtext[1]{{\color{black}{#1}}}

\journal{Information Systems}

\begin{document}
\begin{frontmatter}



\title{Learning policies for resource allocation in business processes}


\author[label1]{Jeroen Middelhuis\corref{cor1}}
\author[label1]{Riccardo Lo Bianco}
\author[label2]{Eliran Sherzer}
\author[label1]{Zaharah Bukhsh}
\author[label1]{Ivo Adan}
\author[label1]{Remco Dijkman}

\affiliation[label1]{organization={Eindhoven University of Technology, Department of Industrial Engineering and Innovation Sciences}, country={The Netherlands}}
\affiliation[label2]{organization={Ariel University, Department of Industrial Engineering and Management}, country={Israel}}
\cortext[cor1]{j.middelhuis@tue.nl (J. Middelhuis)}

\begin{abstract}
Efficient allocation of resources to activities is pivotal in executing business processes but remains challenging. While resource allocation methodologies are well-established in domains like manufacturing, their application within business process management remains limited. Existing methods often do not scale well to large processes with numerous activities or optimize across multiple cases. This paper aims to address this gap by proposing two learning-based methods for resource allocation in business processes to minimize the average cycle time of cases. The first method leverages Deep Reinforcement Learning (DRL) to learn policies by allocating resources to activities. The second method is a score-based value function approximation approach, which learns the weights of a set of curated features to prioritize resource assignments. We evaluated the proposed approaches on six distinct business processes with archetypal process flows, referred to as scenarios, and three realistically sized business processes, referred to as composite business processes, which are a combination of the scenarios. We benchmarked our methods against traditional heuristics and existing resource allocation methods. The results show that our methods learn adaptive resource allocation policies that outperform or are competitive with the benchmarks in five out of six scenarios. The DRL approach outperforms all benchmarks in all three composite business processes and finds a policy that is, on average, 12.7\% better than the best-performing benchmark.
\end{abstract}



\begin{keyword}
Resource allocation \sep business process optimization \sep deep reinforcement learning \sep Bayesian optimization


\end{keyword}

\end{frontmatter}


\section{Introduction}
Efficiently allocating resources to activities that must be executed in a business process at a particular moment at run-time is crucial for organizations to achieve operational outcomes. Optimally utilizing resources to their full capacity maximizes their productivity and enhances an organization's profitability and competitiveness. 

In recent years, the main contribution to data-driven optimization of business processes has been seen in Prescriptive Processing Monitoring (PrPM) \cite{kubrak_prescriptive_2021}. This research area has seen a strong increase in attention and aims to make recommendations for a case (i.e., an instance of the process) that lead to the best outcome. For example, several PrPM methods recommend the best resource to perform a specific activity for running cases to minimize the case's completion time \cite{sindhgatta_context-aware_2016,wibisono_--fly_2015} or cycle time \cite{thomas_online_2017}. However, the main limitation of PrPM is that it only optimizes for individual cases and does not consider the effect of intervention for one case on other cases \cite{kubrak_prescriptive_2021}.

\begin{figure}[!h]
    \centering
    \includegraphics[width=0.6\linewidth]{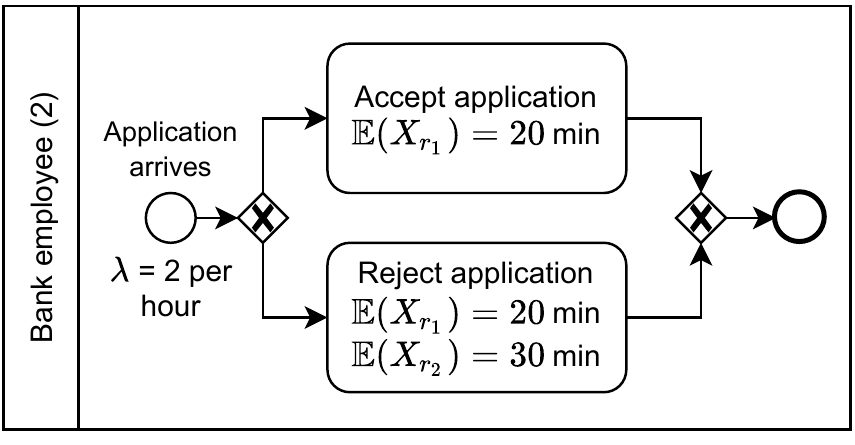}
    \caption{An example of a business process model of a loan application process with two activities and two resources. Cases arrive according to some distribution with rate $\lambda$, and the loans are either accepted (AA) or rejected (RA). Each activity can be performed by both resources $r_1$ and $r_2$, according to a processing time distribution with mean $\mathbb{E}(X_{i})$.}\label{fig:model}
\end{figure}

Data-driven business process optimization (BPO) methods aim to optimize the performance of the complete process and not just individual cases (e.g., ~\cite{park_prediction-based_2019, huang_reinforcement_2011, firouzian_cycle_2019, zbikowski_deep_2023}). In BPO, operational decisions are made to optimize process key performance indicator(s) of interest, such as the mean cycle time or throughput, instead of single case outcome. Take as an example the business process in Figure~\ref{fig:model}. This process has two activities, \textit{Accept Application (AA)} and \textit{Reject Application (RA)}, and two resources, $r_1$ and $r_2$. Activity \textit{AA} can only be performed by resource $r_1$, and the activity \textit{RA} can be performed by both resources. In a situation where two cases are present, case \textit{A} at activity \textit{AA} and case \textit{B} at activity \textit{RA}, a PrPM method would recommend using resource $r_1$ for both cases as it the most efficient resource. However, assigning resource $r_1$ to case \textit{B} means that case \textit{A} needs to wait until resource $r_1$ becomes available again. From a process perspective, assigning resource $r_2$ to case \textit{B} and resource $r_1$ to case \textit{A} would be efficient as both cases can be processed simultaneously and are not competing with each other over the available resources.


The online resource allocation problem in business processes is a dynamic task assignment problem \cite{spivey_dynamic_2004}. Several approaches have been proposed to solve this problem. Park and Song \cite{park_prediction-based_2019} predicted the next activities and processing times of activity-resource pairs in running cases and solved a min-cost max-flow network problem to find optimal resource assignments. Huang et al. \cite{huang_reinforcement_2011} used Reinforcement Learning (RL) to learn the effective resource assignments. This approach was extended by Firouzian et al. \cite{firouzian_cycle_2019} by including entropy-based features mined from the event log. However, the tabular Q-learning method used in these papers is unsuitable for problems with large state spaces, such as business processes containing numerous activities and resources. Žbiwkowski et al. \cite{zbikowski_deep_2023} partly solved this problem using Deep Reinforcement Learning (DRL), where a neural network replaces the Q-table to maximize the number of completed cases. Their method, however, learns feasible actions through experience, which becomes computationally expensive in larger business processes and may not converge to competitive policies.

In this paper, we present two learning methods for resource allocation in business processes to minimize the mean cycle time of cases. The first method leverages DRL to learn effective policies by interacting with the simulator of a business process. It can autonomously derive high-dimensional features directly from the state information. The second method is a score-based value function approximation (SVFA) method, which leverages domain expertise to curate process features and learns weights for these features to rank different resource assignments.  

To evaluate our models, we first created six distinct business process models, referred to as scenarios, with archetypal business process flows and characteristics, such as parallel and choice patterns, varying resource utilization rates, and resource competition. Subsequently, these scenarios were combined in sequence, in sequence with reversed order, and in parallel to create three realistically sized business processes. We benchmarked our methods with traditional heuristics and existing BPO methods \cite{park_prediction-based_2019, zbikowski_deep_2023}. The results show that our learning methods outperform or are competitive with the benchmarks in five out of the six business processes. Our DRL method outperforms all of the benchmarks in all three of the composite business processes.

The outline of this paper is as follows. Section~\ref{sec:related work} gives an overview of the related work. Section~\ref{sec:preliminaries} introduces definitions and concepts related to the execution of business process models. Section~\ref{sec:method} presents our two learning methods to solve the resource allocation problem. Section~\ref{sec:evaluation} introduces the business processes used for evaluation and details the training and evaluation protocol. Section~\ref{sec:conclusion} concludes our work.
\section{Related work}\label{sec:related work}
In business process management, much research has been done on the behavior and allocation of resources. Several methods study the behavior of resources resulting from the current resource allocation policy. This behavior can be evaluated using a set of resource behavior indicators, such as resource preference and productivity \cite{pika_extensible_2014}, or from different perspectives \cite{huang_resource_2012}. López-Pintado et al. \cite{lopez-pintado_prosimos_2023} developed specialized tools for discovering resource profiles to differentiate resource behavior within roles. Furthermore, the current resource allocation policy may also be used to assess the efficiency of a business process \cite{duran_analysis_2019}. 

While these methods generate valuable insights into the current resource allocation policy, they do not consider optimizing existing policies. Our work focuses on improving the resource allocation policy itself, and therefore, we will discuss studies that propose algorithms for resource allocation. We can distinguish between rule-based and dynamic resource allocation methods in the literature.

Several rule-based methods use process mining techniques to determine static resource allocation strategies. Xu et al. \cite{xu_resource_2008} assigned resources to activities based on the workflow of the process model. Other methods use resource-related information mined from the event log in various dimensions to create a ranking of resource assignments, which is used to assign resources to activities at run-time \cite{zhao_optimization_2015, arias_framework_2016, kuchar_automatic_2016}. A generic genetic algorithm was also used for a resource assignment scheme for a specific process based on colored Petri nets \cite{si_petri_2018}. These methods provide a generalized resource allocation scheme for the business process model but are not adaptive to the specific circumstances of the process, such as the number of available tasks and resources.

Dynamic methods make assignments based on the current circumstances of the process. Park and Song \cite{park_prediction-based_2019} proposed a scheduling approach that uses an offline prediction model to predict the next activities and processing times of resource assignments of a running case. These predictions serve as input for the minimum cost and maximum flow algorithm, which finds optimal resource-activity pairs at each decision step. However, the method only optimizes for the current decision step and does not consider the long-term effects of decisions.

In recent years, RL has become increasingly popular and has also been applied to business process optimization. RL is an optimization method for sequential decision-making problems, such as resource allocation, where a decision-making agent learns a policy by taking actions in an environment \cite{sutton_reinforcement_2018}. At each decision step, the agent receives specific information from the business process called the state and, based on the action taken, receives a reward signal. Huang et al. \cite{huang_reinforcement_2011} proposed RLRAM, a method that uses the RL algorithm Q-learning, where a Q-table stores the expected future rewards for each state-action pair. They use action-dependent features in their state representation and train their agent to minimize the flow time of the process. This method was extended by Firouzian et al. \cite{firouzian_cycle_2019}, who added entropy-based features mined from the event log. While this method works relatively well for problems with small state spaces, the method becomes unsuitable for processes with large state spaces, as separate Q-values are stored for each state. \.Zibowski et al. \cite{zbikowski_deep_2023} solved this issue by DRL to maximize the number of completed cases in a business process. The DRL approach replaces the Q-table with a neural network approximating the value function. However, their method must learn which actions are feasible in each state, which is computationally expensive and may not converge to competitive policies.

Manufacturing processes are closely related to business processes, and numerous methods have been proposed for resource allocation in manufacturing. Many heuristics exist, such as Earliest Due Date or First-In-First-Out \cite{sels_comparison_2012}. While these heuristics are robust, explainable, and computationally efficient, they are less effective in situations for which they are not explicitly developed \cite{nie_gep-based_2013}. DRL has also been applied in job shop scheduling~\cite{cunha_deep_2020}. For example, Luo \cite{luo_dynamic_2020} trained an agent to select dispatching rules, and Song et al. \cite{song_flexible_2023} solved a dynamic assignment problem by employing graph neural networks and DRL to capture complex structural relationships between different resources and jobs. However, these methods often assume the sequence of activities in a case is known, which is not true for business processes.

In this paper, we present two dynamic methods that address the limitations of existing methods for resource allocation in business processes. \btext{Our DRL method extends current research on DRL~\cite{zbikowski_deep_2023} by improving scalability and efficiency while optimizing a more complex objective. Our SVFA approach, which employs a ranking of assignments similar to the methods in \cite{zhao_optimization_2015, arias_framework_2016, kuchar_automatic_2016}, improves these techniques by learning optimal feature weights and dynamically adjusting the ranking to the state of the business process.} Our methods are adaptive to the current circumstances of the process and can learn near-optimal policies for realistically sized business processes.
\section{Preliminaries}\label{sec:preliminaries}
\btext{
This section introduces definitions and concepts which we will use in this paper. Section~\ref{sec:business process models} provides essential definitions for understanding business process models, Section~\ref{sec:business process execution} defines how business processes are executed based on their model, and Section~\ref{sec:resource allocation problem} formalizes the problem of allocating resources to activities in a business process.

\subsection{Business process models} \label{sec:business process models}
This paper considers business process models describing the activities that can be performed in an organization and the resources that can perform them, including the arrival process of new cases and the processing times of activities (e.g., the business process in Figure~\ref{fig:model}). We assume the process model can be simulated or executed according to some execution semantics, which makes the business process observable. Specifically, at any point in time, we can observe the execution state of the business process and events that change the state of the process, such as the completion of activities or the arrival of new cases. We formalize these aspects using the following definitions.

Let $\mathcal{A}$ be the set of activities in a business process, $\mathcal{C}$ the set of cases,  $\mathcal{T}$ the time domain, $\mathcal{R}$ the set of resources, and $\{start, complete\}$ be the set of activity life cycle transitions that indicate starting and completing an activity.

\begin{definition}[\textbf{Event}]
 $\mathcal{E} = \mathcal{A} \times \mathcal{C} \times \mathcal{T} \times \mathcal{R} \times \{start, complete\}$ is the set of all possible events. An event is a tuple $(a,c,t,r,l) \in \mathcal{E}$.  
\end{definition}

In line with van der Aalst \cite{van_der_aalst_process_2016}, we refer to the different elements (activity, case, time, resource, and life cycle transition) of an event $e$ as $\#_a(e)$, $\#_c(e)$, $\#_t(e)$, $\#_r(e)$ and $\#_l(e)$.
A trace is a finite, non-empty sequence of events that describes everything that happened in a business case.
 
\begin{definition}[\textbf{Trace}]
For $e_1, e_2, \ldots,e_n\in\mathcal{E}$, $\langle e_1,e_2,\dots,e_n\rangle$ is a trace with length $n$. Events in a trace must be non-decreasing in time: $\forall_{1 \leq i\leq j \leq n} \#_t(e_i) \leq \#_t(e_j)$. The length of a trace $\sigma$ is also denoted as $|\sigma|$.
\end{definition}

The process model in Figure~\ref{fig:model} consists of two activities. Activities can be executed multiple times. We refer to the occurrence of an activity as an activity instance. Note that, while in the example of Figure~\ref{fig:model} an activity is executed only once for each case, activities can be executed multiple times for the same case.

\begin{definition}[\textbf{Activity instance}]\label{def:activity instance}
Let $\mathcal{K}$ be the set of all activity instances. $\mathcal{K}_{c}$ is the set of all activity instances of case $c\in\mathcal{C}$ and $\mathcal{K}_{a}$ is set of all activity instances of activity $a \in \mathcal{A}$. An activity instance $k\in\mathcal{K}$ has a life cycle state $\#_p(k) \in \{waiting, processing, complete\}$.
\end{definition}

We also use subscripts to indicate to what case $K_c\in\mathcal{K}_c$ and activity $K_a\in\mathcal{K}_a$ an activity instance belongs.

To model the execution state of a business process for the purpose of learning what the best resource allocation is in a particular state, we must be able to observe what activity instances are waiting for processing and what resources are available to be assigned to these activity instances. An unassigned activity instance is an activity waiting in the queue of an activity to be executed by a resource. 

\begin{definition}[\textbf{Unassigned Activity Instance}] \label{def:unassigned instance}
An unassigned activity instance is an element from the set $\{k\in\mathcal{K}\mid\#_p(k)=waiting\}$.
\end{definition}

An available resource is a resource that is currently not executing an activity instance and therefore can be assigned to execute one.

\begin{definition}[\textbf{Available Resource, Unavailable Resource}] \label{def:availability}
$R^+ \subseteq \mathcal{R}$ is the set of available resources and $R^- \subseteq \mathcal{R}$ the set of unavailable resources, such that $R^+ \cup R^- = \mathcal{R}$ and $R^+ \cap R^- = \emptyset$.
\end{definition}

A resource is removed from $R^+$ and added to $R^-$ when it starts executing an activity instance. A resource is moved from  $R^-$ to $R^+$ when it finishes processing the activity instance to which it was assigned.

A business process defines which resource is eligible to perform which activity. Not every resource can perform each activity due to, for example, a lack of a specific skill set or level of authorization. 

\begin{definition}[\textbf{Resource Eligibility}] \label{def:eligibility}
For an activity $a \in \mathcal{A}$, $\mathcal{R}_a \subseteq \mathcal{R}$ is the set of resources that can execute $a$, which implies that $r \in \mathcal{R}_a$ can also execute each $k \in \mathcal{K}_a$.
\end{definition}

For example, the process model from Figure~\ref{fig:model} defines that both resources can perform activity \textit{RA}, while only resource $r_1$ can perform activity \textit{AA}. Therefore, the eligible resource set for activity \textit{RA} is  $\mathcal{R}_{RA}=\{r_1,r_2\}$, and for activity \textit{AA} is $\mathcal{R}_{AA}=\{r_1\}$.

\subsection{Business Process Execution}\label{sec:business process execution}
The way in which business processes can be executed is typically defined in terms of how the state of a business process can transition, depending on what happens in the business process. The way in which business processes can transition in state has been well-defined for different kinds of business process modeling notations, such as BPMN and Petri nets. Accordingly, we assume that a business process model exists that determines when a new case can arrive and which activity (instance) can be executed when a new case arrives or when the execution of another activity instance completes. We extend this with the effect that resource assignment has on state transitions.

The business process execution state contains all information about the condition of the process at a given point in time, such as what resources and activity instances are available, as well as what is currently being processed. From a business process execution state, the process may transition into another state after some time $t$ has elapsed as the consequence of an event occurring. These events can be the arrival of a new case, the assignment of an activity instance, the completion of an activity instance, or the completion of a case. Events happen according to the process model. An activity can only be executed if assigned to a resource.

\begin{definition}[Business Process Execution State]\label{def:state}
Let $C \subseteq \mathcal{C}$ be a set of cases, $K \subseteq \mathcal{K}$ be a set of unassigned activity instances, $R^+, R^- \subseteq \mathcal{R}$ be sets of assigned and unassigned resources, $B \subseteq Z \times R^-$, where $Z = \{k\in\mathcal{K}\mid\#_p(k)=processing\}$, be a current assignment of resources to activity instances, and $t\in\mathcal{T}$ a moment in time, the execution state of a business process is a tuple $(C, K, R^+, R^-, B, t)$ consisting of those elements. $\mathcal{S}$ is the set of all business process execution states.
\end{definition}

In a particular execution state of a business process it is possible to assign resources to activity instances, if they are available and eligible to execute those activity instances.

\begin{definition}[\textbf{Assignment}] \label{def:assignment}
An assignment is a tuple $(r, k)$ of a resource $r \in \mathcal{R}$ and an activity instance $k \in \mathcal{K}$. Considering resource eligibility, the set of eligible assignments $\mathcal{D}=\{(r, k) \mid \forall a \in \mathcal{A}, r \in \mathcal{R}_a, k \in \mathcal{K}_a\}$. The set of possible assignments in a specific business process execution state as $D=\{(r, k) \mid (r,k) \in \mathcal{D}, r \in R^+, k \in K\}$.
\end{definition}

When an assignment is made, the process transitions into a new execution state. This transition occurs due to three things happening: (1) the assignment itself; (2) the completion of the execution of an activity instance; or (3) the arrival of a new case. For (2) and (3), we assume that a business process model exists that describes when cases arrive and which activities can be executed when a case arrives or when another activity completes.

\begin{definition}[\textbf{State Transition}] \label{def:state transition}
From a business process execution state $(C, K, R^+, R^-, B, t) \in \mathcal{S}$, a business process can transition into a new state due to three possible events:
\begin{compactenum}
    \item By assigning $(r,k)\in D$.
    \item By the completion of an activity instance $k\in K$.
    \item By the arrival of a new case $c \notin C$.
\end{compactenum}
When assigning $(r,k)\in D$, the business process transitions into a state $(C, K', R^{+\prime}, R^{-\prime}, B', t)$, where $K'=K-\{k\}$, $R^{+\prime}=R^{+} - \{r\}$, $R^{-\prime}=R^{-} \cup \{r\}$, and $B'=B\cup \{(r,k)\}$.
When $k\in K$ completes with $(r,k)\in B$, the business process transitions into a state $(C, K', R^{+\prime}, R^{-\prime}, B', t)$, where $K'=K-\{k\}$, $R^{+\prime}=R^{+} \cup \{r\}$, $R^{-\prime}=R^{-} - \{r\}$ and $B'=B - \{(r,k)\}$. Subsequently, according to the business process model, new activity instances may be generated and added to $K'$. At any moment, according to the business process model, new cases may arrive, which trigger the addition of a case to the set of cases $C$ and initial activity instances to the set of activity instances $K$.
\end{definition}

\begin{definition}[\textbf{State Transition Function}] \label{def:state transition function}
$\tau: \mathcal{S} \times D \rightarrow \mathcal{S}$ is a transition function that, for a state $S$ and an assignment $(r,k)$ of a resource $r$ to an activity instance $k$ returns the next state $S'$ that is derived through Def.~\ref{def:state transition} and simulation of the business process model. 
\end{definition}

\subsection{Resource allocation problem} \label{sec:resource allocation problem}
The resource allocation problem we aim to solve in this paper is a dynamic task assignment problem \cite{spivey_dynamic_2004}. The solution to such a problem is a policy that, given a business process execution state, assigns a resource to an unassigned activity instance.

We can now define a resource allocation policy as follows.

\begin{definition}[Resource Allocation Policy]\label{def:policy}
A resource allocation policy is a function $\pi: \mathcal{S} \rightarrow \mathcal{R} \times \mathcal{A}$, which for a state $(C, K, R^+, R^-, B, t) \in\mathcal{S}$ returns a possible assignment $(r,k)\in R^+ \times K$.
\end{definition}

The goal is to find an optimal policy. This is the policy that, when applied to the simulator for a particular duration, optimizes a specific objective. The duration can be flexibly chosen, for example, as a particular duration of simulation time or a particular number of cases. We define the duration by setting a function $done: \mathcal{S} \rightarrow Bool$ that returns true if and only if a business process execution state is a final state.

\begin{definition}[Rollout]\label{def:rollout}
Let $S$ be the initial state of a business process, $\pi$ be a policy, $\tau$ be a transition function, and $done$ be the simulation duration function. A rollout $\hat{\sigma}$ is a trace $\sigma$ that is generated by applying $S' = \tau(S, \pi(S))$ until $done(S)$. 
\end{definition}

It is important to see that due to randomness in the business process simulation, for example, in the arrivals and processing times, applying the same policy to the same start state may lead to different rollouts. We define the function that determines the value of a trace.

\begin{definition}[Reward, Value]\label{def:reward} 
Let $\hat{\sigma}$ be a rollout. A reward function is a function $r: \mathcal{E} \rightarrow \mathbb{R}$ that assigns a reward to each event that happens in the rollout. The value of a rollout $\hat{\sigma}$ is
\[V(\hat{\sigma}) = \sum_{i=1}^{|\hat{\sigma}|} r(e)\]

Let $\Sigma$ be a multiset of rollouts that is generated by simulation. The expected value of the policy $\pi$ that generated the rollouts
\[\hat{V}_\pi(\Sigma) = \frac{1}{|\Sigma|} \sum_{\hat{\sigma}\in\Sigma} V(\sigma)\]
\end{definition}

Note that the reward function depends on the optimization objective. In this paper, we aim to minimize the average cycle time of cases in the process.

\begin{definition}[\textbf{Cycle Time Reward}] \label{def:cycle time} 
For a sequence of events of a case $\sigma=\langle e_1,e_2,\dots,e_n\rangle$, the cycle time $c_{CT}$ is the difference between the time of occurrence of the first and the last event, $c_{CT} = \#_t(e_n) - \#_t(e_1)$.
\end{definition}

As the aim is to minimize the cycle time, a policy is optimal if it produces the lowest expected value.

\begin{definition}[Optimal Resource Allocation Policy]\label{def:optimal policy}
A policy $\pi$ is considered to be optimal if there is no other policy $\pi'$ for which $\hat{V}_{\pi'}(\Sigma) < \hat{V}_\pi(\Sigma)$ for a particular set of rollouts.
\end{definition}

In practice, it is hard to determine if there exists no policy with a lower value. In addition to that, the policy that is selected to be optimal depends on the specific set of rollouts for which the values are determined. For that reason, the best policy is typically determined by approximation and subsequent comparison of the policy to other policies.
}
\section{Method} \label{sec:method}
\btext{In this section, we present two learning-based methods for resource allocation in business processes to approximate the optimal resource allocation policy that minimizes the average cycle time of cases (Def.~\ref{def:optimal policy}). First, we present a Deep Reinforcement Learning (DRL) method, and second, we present a Score-based Value Function Approximation (SVFA) method.}

The main differences between the two methods are how each algorithm is trained and how information is represented in each algorithm. The DRL method learns continuously during the simulation of the business process based on a reward it receives after each decision step, whereas SVFA receives a single reward signal at the end of the simulation. SVFA's advantage is that it directly minimizes the objective function, \btext{while in DRL, an agent optimizes a feedback signal from a reward function. The challenge lies in developing a reward function that encourages the agent to optimize the objective.}

The information provided in the SVFA method is represented in a set of hand-crafted features based on domain knowledge and optimizes these weights. On the other hand, the DRL agent uses a neural network that automatically learns important features from the state space. The hand-crafted features contain important information for resource allocation but may overlook vital information hidden in the data, which a neural network can discover.

Section~\ref{sec:method 1} details the DRL method and Section~\ref{sec:method 2} the SVFA method.

\subsection{Deep Reinforcement Learning}\label{sec:method 1}
\begin{figure}[h]
\centering
    \includegraphics[width=0.8\linewidth]{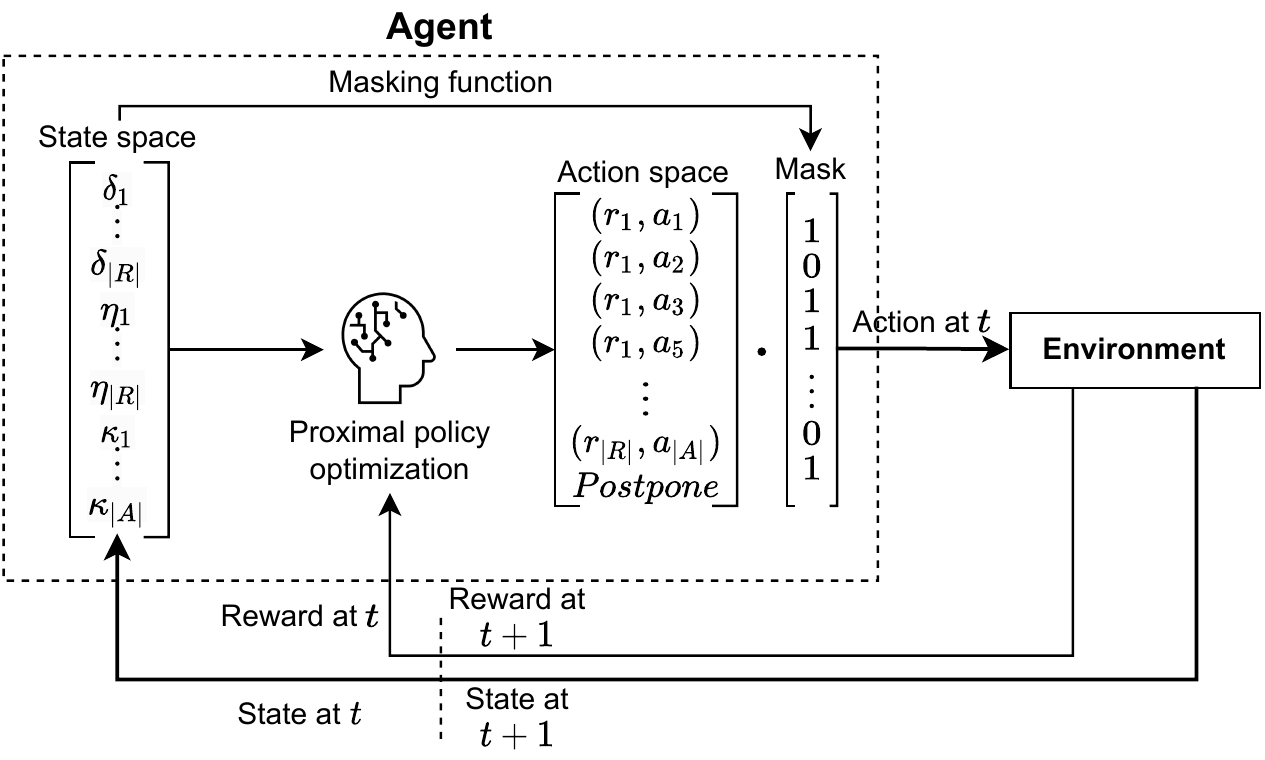}
    \caption{Agent’s perspective of the agent-environment interaction, adapted from \cite{sutton_reinforcement_2018}.}
    \label{fig:RL loop}
\end{figure}

According to Sutton \& Barto \cite{sutton_reinforcement_2018}, RL is a method to train one or multiple agents to interact with an environment to maximize a reward signal. The term DRL is used for RL algorithms that employ neural networks as function approximations. Figure~\ref{fig:RL loop} shows how we developed the reinforcement cycle for resource allocation in business processes. The environment is a discrete-event simulation of the business process. Each time an assignment is possible, according to Def.~\ref{def:assignment}, the simulation prompts the agent to make a decision, which is an assignment. We refer to this moment as a decision step. The environment processes the agent's chosen action, and based on the action's effect on the environment, a reward is returned to the agent (Def.~\ref{def:reward}). One run of a simulation, $\Sigma$, is called an episode and can contain a different number of decision steps, depending on the stochastic events in each rollout. The agent learns a policy, which is a function that maps the state to the actions and defines which action should be taken in a given state to maximize the reward.

In this paper, we use the state-of-the-art DRL algorithm Proximal Policy Optimization (PPO) \cite{schulman_proximal_2017}. Depending on the available resources and unassigned activity instances, several assignments may not be possible at a specific decision step (Def.~\ref{def:assignment}). Therefore, we use a masking function that, given a state, prevents the agent from taking these infeasible actions, improving training speed and scalability \cite{huang_closer_2022}. 

\btext{The main difference between our DRL method and the most relevant previous work on DRL \cite{zbikowski_deep_2023} is that we optimize for a more complex objective. Instead of only maximizing the number of completed cases, we also consider the cycle time in which they are completed. Furthermore, we mask infeasible actions, which significantly improve training speed. For example, in our largest process, there are 24 possible actions. If only one assignment is possible, \cite{zbikowski_deep_2023} must take each action multiple times for every state to learn the feasible actions, while our method receives this information through the masking function. In the former method, there is still no guarantee infeasible actions cannot be taken. Lastly, we use PPO for our problem because it provides a stable and efficient learning process with gradual policy adjustments, which are crucial for handling the dynamic environment of online resource allocation.}

We model the resource allocation problem as a finite Markov Decision Process (MDP) within the DRL framework. The remainder of this section defines each component of the MDP, as shown in Fig~\ref{fig:RL loop}. 
\newline

\noindent \textbf{State} \\
\indent The business process execution state (Def.~\ref{def:state}) contains all information related to the current condition of the process. The state, as defined in the RL paradigm, describes what the agent observes from the business process execution state. To create the features in the state, we order the resources $r \in \mathcal{R}$ and activities $a \in \mathbf{A}$. We use subscript $r_i$ and $a_j$ to indicate the position of a specific resource and activity in the list, where $i \in \{1, \dots, |\mathcal{R}|\}$ and $j \in \{1, \dots, |\mathcal{A}|\}$. The state space contains:
\begin{compactitem}
\item[-] A mapping $\delta_i \rightarrow \{0, 1\}$, indicating for each resource whether it is assigned or not (Def.~\ref{def:availability}), e.g., for resource $r_i : 1$ if $r_i \in R^+$, or $0$ if $r_i \in R^-$.
\item[-] A mapping $\eta_i \rightarrow [0, 1]$, indicating for each resource which activity the resource is assigned to (Def.~\ref{def:assignment}), e.g., for resource $r_i: \{\frac{j}{|\mathcal{A}|} \mid k \in K_{a_j}\}$, if $(r_i, k) \in B$, or 0 otherwise.
\item[-] A mapping $\kappa_j \rightarrow [0, 1]$, indicating the queue of unassigned activity instances of each activity. To reduce the dimensionality of the state space, the queue is normalized by dividing the number of unassigned activity instances by 100 and truncating values higher than 1, i.e.,  for activity $a :\min(\frac{|K_{a}|}{100}, 1)$.
\end{compactitem}
The input of the agent is the state vector. As shown above, the features in the vector are normalized between 0 and 1, which is a common practice to improve sample efficiency during training \cite{henderson_deep_2018}. Given a state, the agent receives information about which resources are available, how many unassigned activity instances are queued at each activity, and the current assignments. Using this information, the agent can learn what assignments lead to high rewards based on the state. Furthermore, it can learn when other resources are likely to be available again, and it can, for example, postpone a decision to wait for a better assignment. The size of the state vector is $2 \cdot |\mathcal{R}| + |\mathcal{A}|$.
\newline

\noindent \textbf{Action} \\
\indent The actions are all assignments (Def.~\ref{def:assignment}) and a postpone action, which delays the current decision step and evolves the environment to the next decision step. At each decision step, a single action is taken. As mentioned, we mask infeasible actions ($\mathcal{D} - D$ according to Def.~\ref{def:assignment}). The masking function is a vector that maps $\mathcal{D} \rightarrow \{0, 1\}$, indicating if an assignment is possible or not, i.e., $1$ if $(r, k) \in D$ and $0$ otherwise. The postpone action is always feasible. The action vector is multiplied by the masking vector, setting the probability of choosing infeasible actions to zero. The size of the action vector is $|\mathcal{D}| + 1$.
\newline

\noindent \textbf{State transition} \\
\indent Given a specific state and action, the business process transitions to a new state, according to Def.~\ref{def:state transition}. Multiple actions can be taken at the same moment in time, depending on the availability of resources or unassigned activity instances. In this case, actions are taken one at a time until no assignments are possible or the agent chooses the postpone action. If there are no possible assignments, the environment evolves until the next decision step. If the agent postpones, the environment evolves until there is at least one different unassigned activity instance or available resource compared to the previous state.
\newline

\noindent \textbf{Reward} \\
\indent The optimization objective is to minimize the average cycle time of cases (Def.~\ref{def:cycle time}), and we should reward the agent for actions that contribute to this objective. \btext{As the agent maximizes cumulative discounted rewards, we inverse the cycle time reward (i.e., Def.~\ref{def:reward}) and give a reward of $\frac{1}{c_{CT}+1}$ for each case $c$ that is completed.} We inverse the cycle time such that a low cycle time results in a high reward and add $+1$ to the denominator \btext{to ensure the rewards do not explode. The reward of a case completion is credited to the action that triggered the state transition in which the case completion happened. Other actions receive a reward of 0.}

Having defined these elements, we can train an agent that learns a policy function for resource allocation that minimizes the average cycle time of cases. The implementation details of the algorithm are given in Section~\ref{sec:ppo results}.

\subsection{Score-based value function approximation}\label{sec:method 2}
The second method we propose is a score-based value function approximation (SVFA) approach that, based on a set of features, \btext{learns to minimize the value of a policy (Def.~\ref{def:reward})}. We consider all possible assignments $(r, k) \in D$ (Def.~\ref{def:assignment}) in a given business process execution state and compute a score for each of them. The score is a function of a set of features related to assignment $(r,k)$ and learned weights for each feature. The features are hand-crafted based on expert knowledge. The features are: 

\begin{compactitem}
    \item[-] $MeanAssignment(r, k)$: the mean processing time of the assignment.
    \item[-] $VarAssignment(r, k)$: the variance of the processing time of the assignment.
    \item[-] $ActivityRank(r, k)$: we evaluate $MeanAssignment(r, k')$ of resource $r$ for all other unassigned activity instances $k' \in K$ and rank them on the mean processing time, where the lowest mean processing time is ranked the highest. $ActivityRank(r, k) \rightarrow \mathbb{N}$ is the rank of unassigned activity instance $k$.
    \item[-] $ResourceRank(r, k)$: we evaluate $MeanAssignment(r', k)$, for all other available resources $r' \in R^+$ and rank them on the mean processing time, where the lowest mean processing time is ranked the highest. $ResourceRank(r, k) \rightarrow \mathbb{N}$ is the rank of resource $r$. 
    \item[-] $ProbFin(r, k)$: the probability that after an assignment $(r, k)$ case $c$ is completed, where $k \in K_c$.
    \item[-] $QueueLength(k)$, the number of unassigned activity instances for a specific activity $|K_a|$, given that $k \in K_a$.
\end{compactitem}
In this paper, features such as $MeanAssignment(r, k)$, $VarAssignment(r, k)$, and $ProbFin(r, k)$ are determined by the business process model. However, an estimate may also be derived from historical data if available. The score function maps each state-action pair to a real value, which represents the goodness of making a specific assignment in a specific state:
\begin{align} \label{eq:score_method2}
      & Score(r, k) = w_1 \cdot MeanAssignment(r, k)  \\ \nonumber 
      & + w_2 \cdot VarAssignment(r, k) +w_3 \cdot ActivityRank(r, k)  \\ \nonumber
      & + w_4 \cdot ResourceRank(r, k) - w_5 \cdot ProbFin(r, k) \\ \nonumber
      &  -w_6 \cdot QueueLength(k), \textrm{where } w_i \geq 0, \textrm{for } i \in \{1, \dots, 6\} \nonumber
\end{align}

To calculate the score, we aggregate the first four features in Equation~\ref{eq:score_method2} by addition, as a lower value indicates better performance (i.e., lower cycle time). Conversely, higher values for the last two features, $ProbFin(r, k)$ and $QueueLength(k)$, are desirable because they signify efficient case completion and minimized queue lengths, which contribute to a lower average cycle time \cite{hopp2011factory}. Therefore, these features are subtracted when calculating the score.

At each decision \btext{step}, the score for all possible assignments (Def.~\ref{def:assignment}) is calculated. The assignment with the lowest score is the best and will be made unless its score does not exceed a learned threshold $w_7$. Instead, the agent postpones if no assignment's score is lower than this seventh weight. Similarly to the DRL approach, when multiple assignments are possible, the agent makes assignments one at a time and then reevaluates all scores based on the new state to make the next assignment.

We use Bayesian optimization to find weights that minimize the average cycle time. Bayesian optimization is a global optimization technique that uses a probabilistic model to predict the relationship between the input parameters and the output of a function. The algorithm learns as new data is obtained and focuses on selecting parameters in promising areas of the search space. \btext{In our setting, the input parameters are the feature weights, and the output is the mean cycle time after an episode. In contrast to our DRL approach, SVFA does not learn \bbtext{from the decision steps in an episode but updates the policy at the end of the episode, based on the average cycle time following the current policy.}} For more information on the Bayesian optimization algorithm, we refer to Frazier \cite{frazier2018tutorial}. The implementation details are given in Section~\ref{sec:bayes_opt}.
\section{Evaluation}\label{sec:evaluation}
Section~\ref{sec:experimental setup} introduces six business processes with archetypal process flows and characteristics and three realistically sized business processes on which we evaluate our methods. Section~\ref{sec:training setup} describes the implementation details of our methods and the hyperparameters used for training them. Section~\ref{sec:evaluation protocol} details the evaluation protocol and introduces the benchmarks. Section~\ref{sec:results} shows the results of our methods compared with three heuristic benchmarks and two benchmarks from the literature.

\subsection{Experimental setup} \label{sec:experimental setup}
This section presents the business processes on which we evaluate and benchmark our methods. First, we design six different business processes, each consisting of two activities and two resources, to analyze which methods work well in what specific situations. We refer to these six business processes as scenarios. Each scenario is \btext{an abstraction of pattern commonly seen in business processes}. The scenarios are characterized by aspects such as parallel or choice constructs, varying utilization rates, and competition over resources. Second, we connect these six scenarios sequentially and in parallel to create three realistically sized business processes containing all the different process flows seen in the scenarios. Evaluation of these business processes demonstrates which methods perform well under specific circumstances (i.e., the scenarios) and which generalize well to a process in which all scenario dynamics are combined. We refer to these three business processes as composite business processes.

\begin{figure}
     \centering
     \begin{subfigure}[b]{0.49\textwidth}
         \centering
         \includegraphics[width=\textwidth]{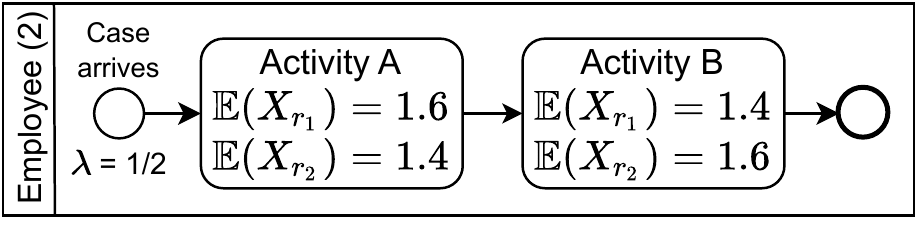}
         \caption{Scenario 1: Low utilization}
         \label{fig:low utilization}
     \end{subfigure}
     \hfill
     \begin{subfigure}[b]{0.49\textwidth}
         \centering
         \includegraphics[width=\textwidth]{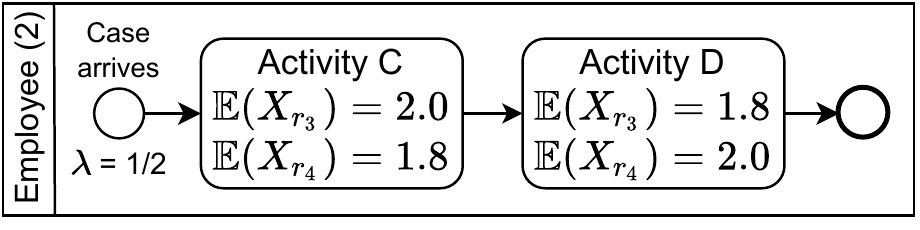}
         \caption{Scenario 2: High utilization}
         \label{fig:high utilization}
     \end{subfigure} \\
     \begin{subfigure}[b]{0.49\textwidth}
         \centering
         \includegraphics[width=\textwidth]{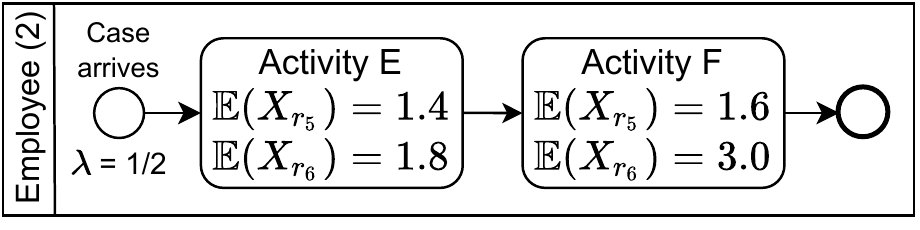}
         \caption{Scenario 3: Slow server}
         \label{fig:slow server}
     \end{subfigure}
     \hfill     
     \begin{subfigure}[b]{0.49\textwidth}
         \centering
         \includegraphics[width=\textwidth]{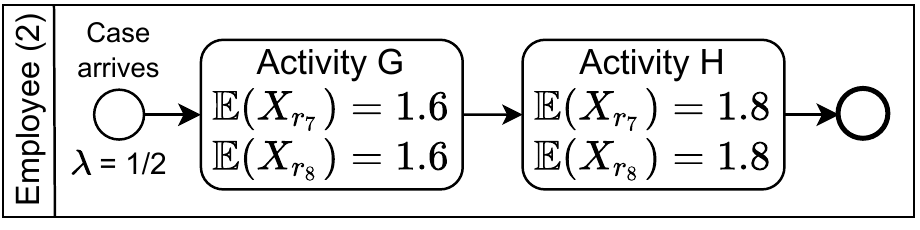}
         \caption{Scenario 4: Slow downstream}
         \label{fig:slow downstream}
     \end{subfigure}\\
     
     \begin{subfigure}[b]{0.49\textwidth}
         \centering
         \includegraphics[width=\textwidth]{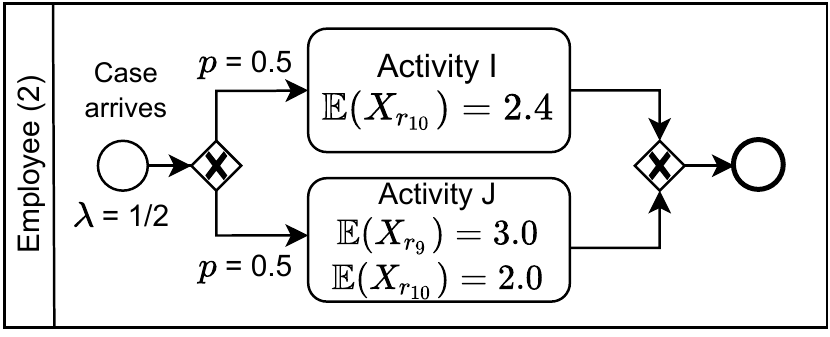}
         \caption{Scenario 5: N-network}
         \label{fig:n-network}
     \end{subfigure}
     \hfill
     \begin{subfigure}[b]{0.49\textwidth}
         \centering
         \includegraphics[width=\textwidth]{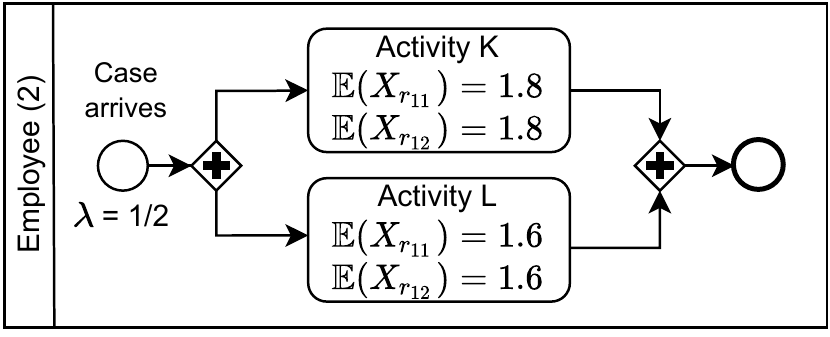}
         \caption{Scenario 6: Parallelism}
         \label{fig:parallism}
     \end{subfigure}
        \caption{Six business process models with different characteristics used for evaluation.}
        \label{fig:scenarios}
\end{figure}

We first introduce the six scenarios. In each scenario, cases arrive according to a Poisson process with rate $\lambda=\frac{1}{2}$ time units. \btext{This arrival rate ensures the system remains stable while still offering a load on the resources such that the problem is challenging. In Section~\ref{sec:results}, we show that the processes become unstable when using higher arrival rates.} The processing times are exponentially distributed with mean $X_r$, $r \in \mathcal{R}$. The mean processing times are chosen to align with the characteristics of their respective scenario (e.g., utilization level, a slower resource) and ensure that each scenario's cycle time is in the same order of magnitude. The latter is important for connecting the scenarios to create the composite business processes such that a single scenario does not dominate the total cycle time of the composite process. The six scenarios are shown in Figure~\ref{fig:scenarios}.

Figure~\ref{fig:low utilization} and Figure~\ref{fig:high utilization} show the first two scenarios characterized by their utilization rates \btext{and represent situations in which the process operates under different loads, such as peak and off-peak hours}. The process's utilization rate significantly impacts the average queue length at an activity $|K_a|$ \cite{hopp2011factory}. If the utilization rate is low, such as in scenario 1 (Figure~\ref{fig:low utilization}), the average size of the queues in the process will also be smaller, and therefore, the agent should focus on minimizing the processing time. Scenario 2 in Figure~\ref{fig:high utilization} has a high utilization rate, which means that the average queue length will be higher, and cases will spend most of their cycle time queuing. Therefore, minimizing queueing time becomes more important in this scenario.

\btext{Figure~\ref{fig:slow server} shows a variant of the slow server problem which is challenging and well-studied in the literature \cite{rubinovitch_slow_1985}. With respect to business process, this scenario contains an experienced employee ($r_5$) who works faster than an inexperienced employee ($r_6$).} An efficient assignment policy should avoid assigning the inexperienced to activity $F$, as this resource is especially slow at executing it. Scenario 4 in Figure~\ref{fig:slow downstream} shows a business process where both resources have the same processing time distribution for each activity. Therefore, the prioritization of activities is more important than the resource that executes them. \btext{This problem is relevant for processes with standardized processing times.}

Figure~\ref{fig:n-network} models scenario 5, a process with two activities with different resource eligibility. This process is a variant of the well-studied bipartite matching problem \cite{cadas_bipartite_2019}. It is also known as the `N-network', referring to the `N' shape of the bipartite graph, where nodes represent resources and activities, and edges represent the resource eligibility (Def.~\ref{def:eligibility}). The exclusive gateway (XOR) characterizes the process, where cases go to either activity with probability $p=0.5$. This scenario represents a situation where one employee ($r_{10}$) can perform more activities due to, for example, the authorization level. This resource's capacity should be balanced over the activities. Figure~\ref{fig:parallism} shows scenario 6, which contains a parallel process flow typically seen in business processes. In this scenario, both activities can be performed simultaneously, but a case is only completed once both activities have been executed. The processing time at each activity is the same for both resources, making prioritizing what activity to complete the most important choice.

In the second part of the evaluation, we trained and tested our models on the three composite processes. The first composite business process is created by sequentially connecting the scenarios, starting with scenario 1 and ending with scenario 6. In the second composite business process, we also connect the scenarios sequentially but in reverse to investigate the impact of the order of the scenarios. Note that in these processes, the external arrival process is still a Poisson process, while the internal arrivals now depend on departures from previous activities. The third composite business process is created by putting all scenarios in parallel.

\bbtext{These scenarios provide typical resource-allocating problems that require adaptive methods to solve them. While these scenarios can be extended by, for example, incorporating additional resources or activities, they provide a good foundation for understanding how different algorithms perform in specific conditions, specifically different control flow patterns, resource loads, and authorization structures (such as the N-model). Moreover, applying our methods to the composite models allows us to study which methods can handle all scenario dynamics simultaneously.}

\subsection{Training setup} \label{sec:training setup}
In this section, we describe the training setup that we used to train our methods. Section~\ref{sec:ppo results} describes the training procedure and the hyperparameters used for the DRL approach. Section~\ref{sec:bayes_opt} describes how SVFA's optimal feature weights are obtained for each business process.

\subsubsection{Deep Reinforcement Learning} \label{sec:ppo results}
We performed hyperparameter tuning for the sequential composite business process using grid search and used the found optimal hyperparameters to train an agent for all business processes considered in this paper. The hyperparameter ranges and optimal values are shown in Table~\ref{tab:hyperparameters}.

\begin{table}[h]
\centering
\caption{Hyperparameter ranges for grid search on the composite business process and optimal hyperparameters found for training maskable PPO agent.}
\label{tab:hyperparameters}
\begin{tabular}{lll}
\toprule
Parameter                  & Range                 & Optimal \\ \midrule
Number of layers           & [1, 2]             & 2      \\
Neurons per layer          & [32, 64, 128]         & 128     \\
Clip range                 & [0.1, 0.2]      & 0.2    \\
Number of samples per update & [25600, 51200]& 25600  \\
Batch size                 & [256, 512]      & 256    \\
Learning rate              & [3e-3, 3e-4, 3e-5]    & 3e-5  \\ 
Discount factor            & [0.99, 0.999, 0.9999] & 0.999 \\ \bottomrule
\end{tabular}
\end{table}

After tuning the hyperparameters, we found that a high number of observations per update and a large batch size significantly improved the model's performance.  We set the number of samples before updating the model to include at least an entire episode. In the composite business processes, there are 11 activities per case, requiring at least the same number of actions before the case is completed and a non-zero reward is returned. When using a larger batch size, more of the delayed and sparse rewards are included in each model update, which gives a more accurate estimate of the gradient and helps reduce the sample noise.

During training, we decrease the learning rate linearly with the total training steps to encourage exploration at the start and exploitation of the policy at the end of training. The discount factor determines the importance of future rewards. We found a high value to be optimal to account for the long-term effects of actions on the cycle time and sparse rewards. We evaluated the current model during training and saved the best model found. We train each model until the policy converges or until a maximum of 2e7 training steps (i.e., decision steps). However, we found that each policy converged much earlier than the maximum number of steps.


We use the maskable PPO implementation of the contributory version of the popular DRL Python library Stable-Baselines3~\footnote{https://github.com/Stable-Baselines-Team/stable-baselines3-contrib} and create our environment using Gymnasium~\footnote{https://gymnasium.farama.org/index.html}. Other hyperparameters were kept at their default values.

\subsubsection{Score-based value function approximation}\label{sec:bayes_opt}
To train the weights of the SVFA method, we use Bayesian optimization. We define a search space in which we optimize the value function's weights for each business process. In our setup, we use $0 \leq w_i \leq 100$, for $i \in \{1, \dots, 7\}$. \btext{The optimization procedure runs for a maximum of \btext{20} trials. Each trial contains 5000 simulations of the business process to estimate the mean cycle time given the current weights.} The method was implemented in Python using the scikit-optimize package\footnote{https://scikit-optimize.github.io/stable/}. 

\subsection{Evaluation protocol and benchmarks} \label{sec:evaluation protocol}
This section describes our approach to evaluate each method and the benchmarks. We first describe how we evaluated the methods and then introduce the benchmarks and their implementation specifics.

Each method is evaluated on 100 simulations of 5000 time units in each business process. The cycle time is calculated according to Def.~\ref{def:cycle time}. Uncompleted cases at the end of the process are truncated and included in the average cycle time calculation. This truncation ensures that the best policy is not always to postpone, as otherwise, no case completions would result in a zero average cycle time. \btext{For each method and benchmark, we will report the mean cycle time and the 95\% confidence interval (CI). A Student's t-test for two independent samples ($p\ge0.05$) will be performed between the best-performing method and the other methods. The best-performing method and methods not significantly different will be reported in bold.}

\btext{We evaluated our methods and will present our results around the proposed arrival rate $\lambda=0.5$. However, to investigate the performance of our models under different loads, we evaluated each method using both different constant arrival rates and a dynamic arrival pattern. We trained and evaluated each model for the arrival rates $\lambda \in \{0.3, 0.35, 0.4, 0.45, 0.5, 0.55, 0.6\}$. In an arrival pattern, the arrival rate changes over time. Figure~\ref{fig:arrival_pattern} shows the arrival pattern used derived from the BPI12W event log.

\begin{figure}[h]
    \centering
    \includegraphics[width=0.7\linewidth]{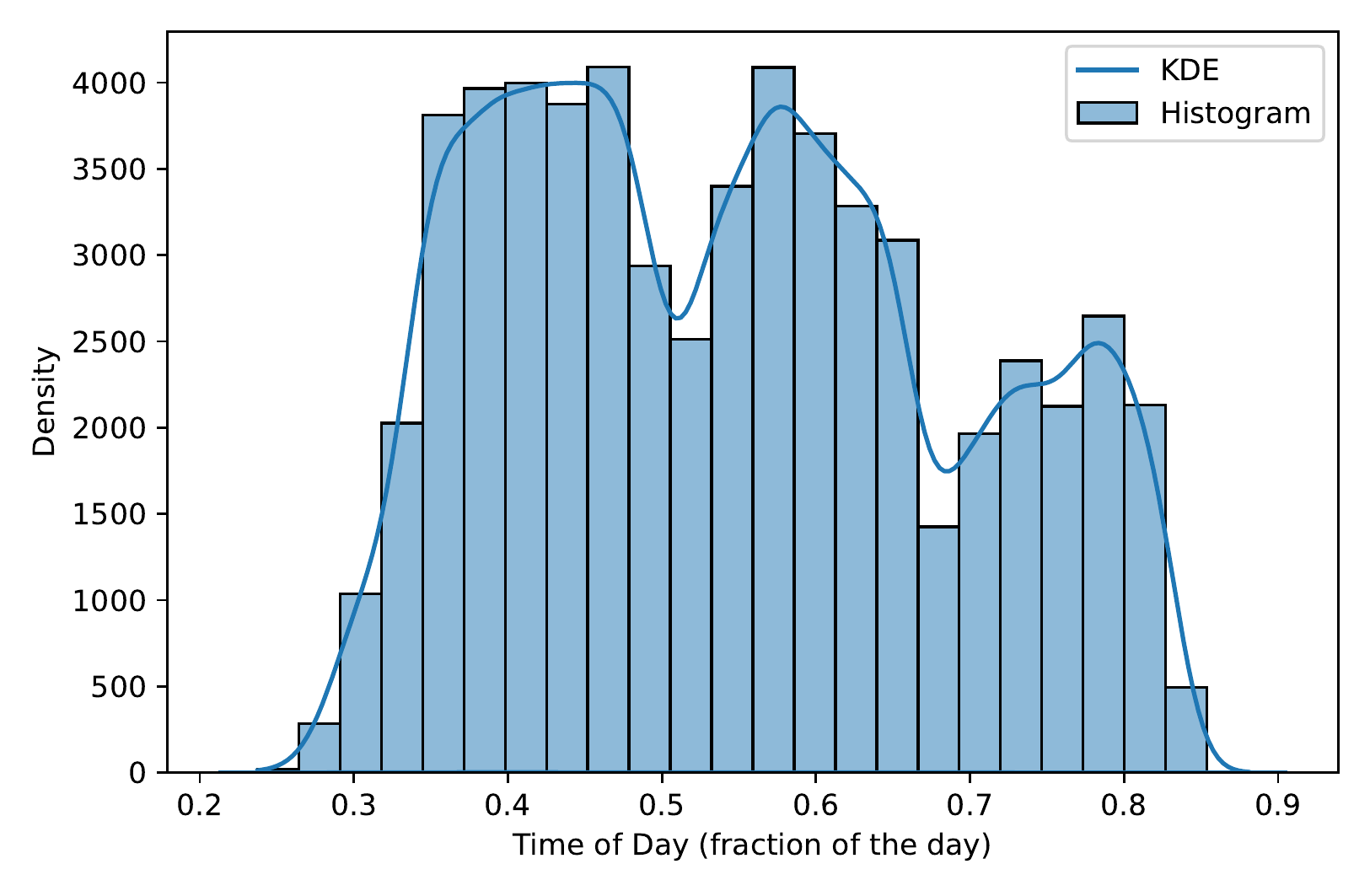}
    \caption{Arrival pattern based on the BPI12W event log}
    \label{fig:arrival_pattern}
\end{figure}

During the simulation, arrivals are sampled using the KDE curve. The pattern repeats every 250-time units, and the night time of the original event log is ignored, as our resources are continuously available. If we include this downtime in arrivals, many cases will arrive during the day and must be processed at night. The pattern is scaled such that the average arrival rate is 0.5. The pattern is implemented as follows. After each arrival, a new arrival is sampled using the maximum arrival rate in the KDE curve ($\lambda_{max}=0.88)$. This arrival is then accepted with a probability proportional to the current arrival rate. For example, if $\lambda_{current} =0.5$, then $P(accept)=\frac{\lambda_{current}}{\lambda_{max}}=0.44$.}

To benchmark our methods, we compare them with three traditional heuristics and two methods from the literature \cite{park_prediction-based_2019, zbikowski_deep_2023}. \bbtext{Some of the considered benchmarks were not directly implementable, or no source code was available. Therefore, we had to make some small changes or assumptions.} The benchmarks and their implementation details are:

\begin{compactitem}
    \item[-] Shortest Processing Time (SPT): all possible assignments at a decision step are evaluated at a decision step, and the assignment with the lowest expected processing time is chosen.
    \item[-] First In, First Out (FIFO): the case that is in the business process for the longest amount of time is prioritized at every activity.
    \item[-] Random: a random assignment is made from all possible assignments at a decision step.
    \item[-] Park and Song \cite{park_prediction-based_2019}: this method uses a prediction model to predict the next activity and the processing time of possible assignments. We do not have an event log to train a prediction model, but can derive the next activity and the expected processing times from the business process model. Therefore, the predictions we passed to the scheduling stage are 100\% accurate.
    \item[-] \.Zbikowski et al. \cite{zbikowski_deep_2023}: \btext{this method uses a double deep Q-network (DDQN) algorithm to train an agent to make assignments. As our business processes are larger than those evaluated in \cite{zbikowski_deep_2023}, we train the DDQN for more steps, equal to the number of steps we train our models. We use the same values for the number of exploration episodes and replay buffer as in the DDQN's implementation.}
\end{compactitem}

We chose these benchmarks because their methods are closely related to our methods and are suitable for the online scheduling problem. Considering we do not need to make any predictions, the minimum-cost maximum-flow algorithm by Park and Song \cite{park_prediction-based_2019} is an advanced version of SPT and, therefore, a relevant benchmark. We implemented RLRAM \cite{huang_reinforcement_2011}; however, the method did \btext{not converge to a competitive policy in any of our business processes}. Our implementation of RLRAM algorithm can be found in our \btext{repository}. Other methods from the literature mine features from an event log to solve the resource allocation problem \cite{firouzian_cycle_2019,xu_resource_2008, zhao_optimization_2015, arias_framework_2016, kuchar_automatic_2016}. However, our evaluation framework uses simulated data without a real-world event log; therefore, these event log-based approaches could not be replicated.

\subsection{Results}\label{sec:results}
\setlength{\tabcolsep}{2pt}

\btext{In this section, we present and discuss the results of our methods compared to the benchmarks. Section~\ref{sec:eval_constant} presents the results of the scenarios and composite business processes using a constant arrival rate of $\lambda=0.5$. Section~\ref{sec:eval_varying} presents the results when using a range of constant arrival rates and a dynamic arrival pattern.

\subsubsection{Evaluation of the business processes}\label{sec:eval_constant}
This section shows the results of our methods compared with the benchmarks for the scenarios and composite business processes. Table~\ref{tab:results scenarios} shows the results of our methods on the six scenarios. The method by \.Zbikowski et al. \cite{zbikowski_deep_2023} can select infeasible actions, after which the state changes and another action is selected. However, in some situations, the state does not change anymore, and the policy gets stuck, selecting infeasible actions. For these scenarios, we do not report the results for this method.}

\begin{table}[h]
\caption{Mean cycle time and CIs for the six scenarios ($\lambda=0.5$).}
\label{tab:results scenarios}
\resizebox{1\linewidth}{!}{
\begin{threeparttable}
\begin{tabular}{lcccccccc}
\toprule
Business process        & DRL                  & SVFA                  & SPT                  & FIFO                 & Random &  Park \& Song \cite{park_prediction-based_2019} & \.Zbikowski et al. \cite{zbikowski_deep_2023} \\ \midrule
        Low utilization & \textbf{5.8 (0.10)} & \textbf{5.9 (0.10)} & \textbf{5.9 (0.09)} & \textbf{6.0 (0.11)} & 6.5 (0.13) & 6.0 (0.09) & -\tnote{a} \\
        High utilization & \textbf{17.9 (0.72)} & 19.4 (0.99) & 19.4 (0.96) & 26.5 (1.86) & 33.2 (3.07) & 19.4 (1.01) & -\tnote{a} \\
        Slow server & \textbf{11.5 (0.33)} & 14.7 (0.45) & 26.6 (1.88) & 20.8 (1.86) & 21.2 (1.25) & 20.5 (1.26) & 58.9 (5.60) \\
        Slow downstream & \textbf{10.0 (0.35)} & \textbf{10.0 (0.32)} & 14.9 (0.61) & \textbf{9.9 (0.32)} & 11.5 (0.39) & 14.7 (0.59) & -\tnote{a} \\
        N-network & 6.8 (0.20)\tnote{*} & \textbf{6.1 (0.11)} & 7.1 (0.21) & \textbf{6.0 (0.12)} & 6.5 (0.15) & 6.3 (0.14) & 6.7 (0.11) \\
        Parallel & 10.8 (0.37) & 12.6 (0.54) & 14.1 (0.6) & \textbf{9.8 (0.35)} & 11.1 (0.49) & 13.7 (0.60) & -\tnote{a} \\ \bottomrule
\end{tabular}
  \begin{tablenotes}
  \item[*]Policy trained with -0.1 penalty for postponing
  \item[a]Learned policy only selected infeasible actions, resulting in indefinite postponement.
  \end{tablenotes}
  \end{threeparttable}
}
\end{table}

Table~\ref{tab:results scenarios} shows that both of our methods outperform or are competitive across benchmarks, except for the scenario involving parallelism. \btext{In the low utilization scenario, strategic decisions like idling or prioritizing case completions become more viable as the resource load is lower}. Our methods can effectively learn these policies. On the other hand, in the high utilization scenario, prioritizing the assignment of activity instances to the fastest available resources is crucial to prevent excessive queue buildup. This strategy aligns closely with the SPT heuristic and with Park and Song's method \cite{park_prediction-based_2019}, as shown in these methods' performance. \btext{Our DRL method found a policy that outperforms the other methods while SVFA's policy was competitive with the best-performing heuristics.}

In the slow server scenario, resource $r_5$ is inefficient in executing activity \textit{E} (Figure~\ref{fig:slow server}), and this assignment should be avoided in most situations. Hence, strategically idling to await a faster resource's availability may be beneficial. Park and Song's method \cite{park_prediction-based_2019} also accommodates strategic idling, which is evident in its improved performance over SPT. Both of our methods successfully learned policies that outperform all benchmarks.

\bbtext{In the slow downstream scenario, the processing time of an activity is equal across resources. Therefore, the emphasis shifts from choosing what resource to allocate to what activity to execute. In fact, in this scenario, it is optimal to always perform the activity \textit{H} if possible, as it completes the case and stops the cycle time. While this assignment is not more efficient on the short-term, it must be completed and thus should be prioritized for case completion. This policy is identical to FIFO's policy, which explains its good performance. Our methods are also capable of learning this policy.}

In the N-network, only resource $r_{10}$ is eligible to execute activity \textit{I} (see Figure~\ref{fig:n-network}); therefore, reserving it for that activity is an effective strategy. While SVFA was able to find a policy that outperforms the other methods, our DRL model struggled. The DRL agent learned only to execute activity \textit{J} and postpone otherwise, leading to high immediate rewards. However, this policy resulted in a large queue building up at activity \textit{I}, resulting in a high average cycle time. After introducing a penalty of $-0.1$ for postponing in this scenario, the agent converged to a better policy.

\bbtext{Similar to scenario 4 (Figure~\ref{fig:slow downstream}), the parallelism scenario contains two activities with identical processing times across the two different activities. As both activities must be executed before a case is completed, it is again optimal to assign available resources to the longest waiting case, which is what the FIFO policy does. In contrast, our methods and other benchmarks do not have explicit information about the previously executed activities of a specific case, meaning that no distinction can be made between new cases and partially completed cases. This lack of information makes finding the optimal policy challenging for this scenario, which is reflected by the results.}

\bbtext{While these scenarios may not capture all possible process patterns or flows that can occur in a business process, they capture the most relevant dynamics in terms of possible control-flow relations, resource loads, and authorization structures. For example, the utilization-based scenarios can represent different resource loads that can change as the process executes. Furthermore, the slow server and N-network emulate different resource authorization structures, and the N-network and parallel scenario model the parallel (AND) and exclusive (XOR) control-flow patterns. While we test our models on these scenarios, there are other scenarios, such as a W-network, which can also be investigated. To illustrate the robustness of our conclusions in other scenarios, we will now extend our evaluation to composite business processes. In addition, in another paper, we applied our method to real-world business processes, containing dynamics such as re-entrant flows and the temporal availability of resources \cite{meneghello24}.}

\begin{figure}[h]
    \centering
    \includegraphics[width=1\linewidth]{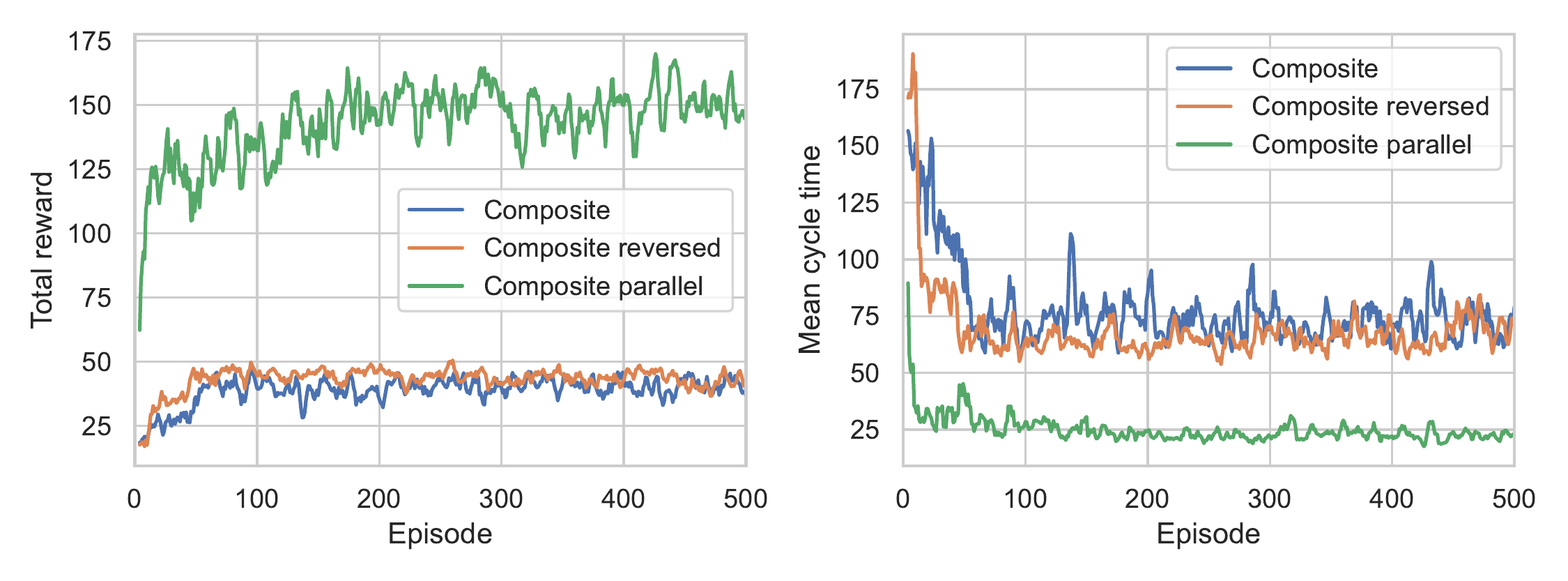}
    \caption{Total reward and mean cycle time during training of DRL models.}
    \label{fig:training_curves}
\end{figure}

\btext{In the second part of the evaluation, we evaluated the three composite business processes. Figure~\ref{fig:training_curves} shows the progression of total reward and mean cycle time per episode during training with our DRL models. The policies for the two sequential processes converge quickly, stabilizing after approximately 100 episodes. The policy for the composite parallel process takes around 200 episodes to converge due to its increased complexity. In this process, various activities can lead to case completion, unlike sequential processes where only the final activity completes the case. All three composite processes converged before reaching the allowed training time (maximum of 2e7 decision steps is approximately 1800 episodes for these processes). In our SVFA method, each model used the 20 allowed trials.}

The results for the composite business processes using a constant arrival rate of $\lambda=0.5 $ are shown in Table~\ref{tab:results complete}.

\begin{table}[h]
\caption{Mean cycle time and CIs for the composite processes ($\lambda=0.5$).}
\label{tab:results complete}
\resizebox{1\linewidth}{!}{
\begin{threeparttable}
\begin{tabular}{lcccccc}
\toprule
Business process        & DRL                  & SVFA                  & SPT                  & FIFO                 & Random  \\ \midrule
        Composite & \textbf{64.7 (2.63)} & 72.9 (2.36) & 100.9 (4.07) & 69.7 (3.5) & 86.5 (4.12) \\
        Composite reversed & \textbf{61.5 (1.50)} & 73.8 (2.48) & 110.7 (4.77) & 70.0 (3.7) & 88.0 (4.53) \\
        Composite parallel & \textbf{23.8 (0.99)} & 33.4 (1.52) & 35.2 (1.71) & 29.3 (1.73) & 41.9 (3.99) \\
\bottomrule
\end{tabular}
\end{threeparttable}
}
\end{table}

Similarly, as seen in some of the scenarios (Table~\ref{tab:results scenarios}), the method by \.Zbikowski et al. \cite{zbikowski_deep_2023} converged to a policy that postponed indefinitely. Furthermore, the inference time of the method by Park and Song \cite{park_prediction-based_2019} was intractable; therefore, we could not evaluate the method for the composite business processes. This limitation was also indicated by the authors. For this reason, the results of these two benchmarks are not included in the evaluation and Table~\ref{tab:results complete}.

\btext{Table~\ref{tab:results complete} shows that DRL consistently outperforms all other methods in composite business processes. The performance of both composite and composite reverse business processes is comparable, indicating that the sequence of scenarios does not significantly affect the outcomes. This result is expected, as the scenarios within the composite business processes remain relatively isolated, each utilizing its own resources. Although FIFO excelled in the parallel scenario, as shown in Table~\ref{tab:results scenarios}, it struggled to handle the complexities of combined scenarios, as shown in Table~\ref{tab:results complete}.

\bbtext{While SVFA performed well in the scenarios, its relative performance decreased in the composite processes. In the two sequential composite processes, SVFA showed similar performance, but in the parallel composite process, SVFA performed slightly worse than FIFO. However, SVFA still outperformed SPT, showcasing that this method is adaptable to larger process instances.} A possible explanation is that there are competing feature weights for each scenario. For example, as can be seen in Table~\ref{tab:results scenarios}, an SPT policy yields good performance in the low utilization scenario, which means that the weight of the feature $MeanAssignment$ should be high for this scenario. On the other hand, SPT makes poor assignments in the slow downstream scenario, which means the weight of $MeanAssignment$ should be low in this scenario. \bbtext{In the composite processes, this feature must be balanced to account for both scenarios, which can result in a sub-optimal policy.}
}

\btext{\subsubsection{Evaluation using varying and dynamic arrival rates}\label{sec:eval_varying}

To further investigate the effectiveness of our methods, we trained and tested them using varying and dynamic arrival rates. First, as discussed in Section~\ref{sec:evaluation protocol}, we trained and evaluated our methods using different constant arrival rates ranging from 0.3 to 0.6. Second, we use an arrival pattern that dynamically changes the arrival rate as the process is executed.

Figure~\ref{fig:varying_scenarios} compares their performance to heuristics in the high utilization and slow server scenario with arrival rates from 0.3 to 0.6. The results show that all policies perform similarly when the process load is low. In these cases, resources are often idle, and simply assigning a resource \bbtext{is already a decent policy. On the other hand, as the load in the process increases, efficiently allocating resources becomes more important. Figure~\ref{fig:varying_scenarios} shows that as the arrival rate increases, the policies diverge. The figure shows that the heuristics are effective for processes with a low load but become unsuitable as the process load and complexity of the assignment problem increase. Our methods, however, are still capable of learning relatively effective policies.}

\begin{figure}[h]
    \centering
    \begin{subfigure}[b]{0.49\textwidth}
        \centering
        \includegraphics[width=\textwidth]{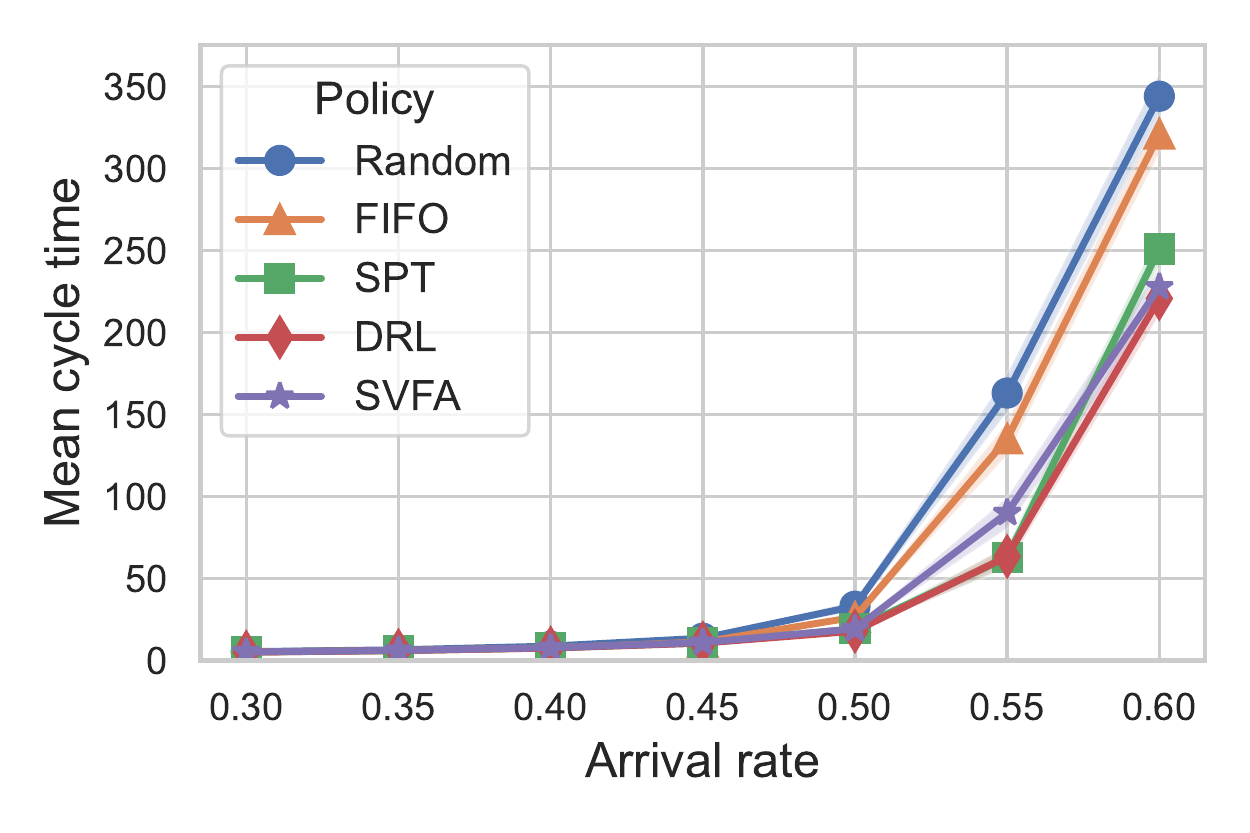}
        \caption{High utilization}
        \label{fig:varying_high_utilization}
    \end{subfigure}
    \hfill
    \begin{subfigure}[b]{0.49\textwidth}
        \centering
        \includegraphics[width=\textwidth]{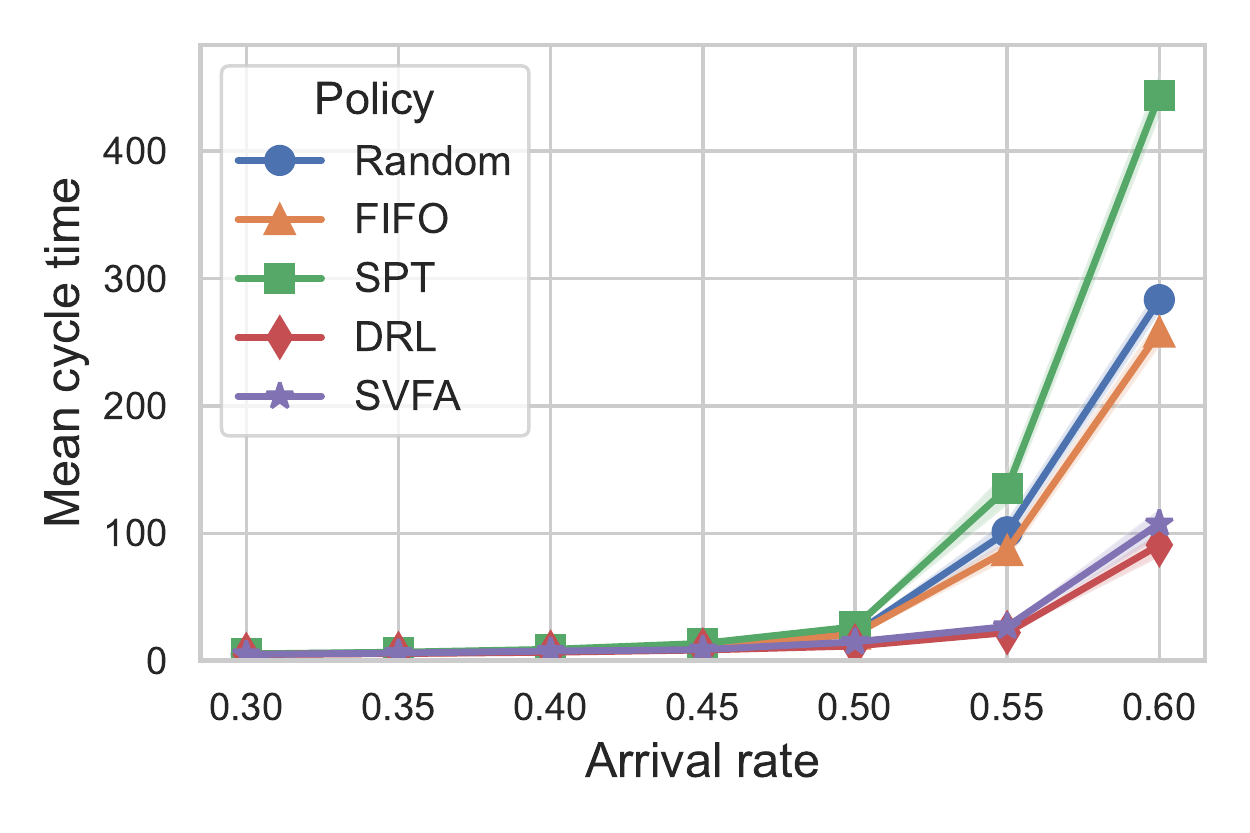}
        \caption{Slow server}
        \label{fig:varying_slow_server}
    \end{subfigure}
    \caption{Comparison of policies in the high utilization and slow server scenarios using a range of arrival rates.}
    \label{fig:varying_scenarios}
\end{figure}

Figure~\ref{fig:varying_scenarios} shows that the cycle time increases rapidly when the arrival rates exceed $\lambda=0.5$, which indicates that the system becomes unstable beyond this point. This pattern is also observable in the utilization rate of the resources. For instance, Figure~\ref{fig:utilization_slow_server} demonstrates the resource utilization in the slow server scenario.

\begin{figure}[h]
    \centering
    \includegraphics[width=1\linewidth]{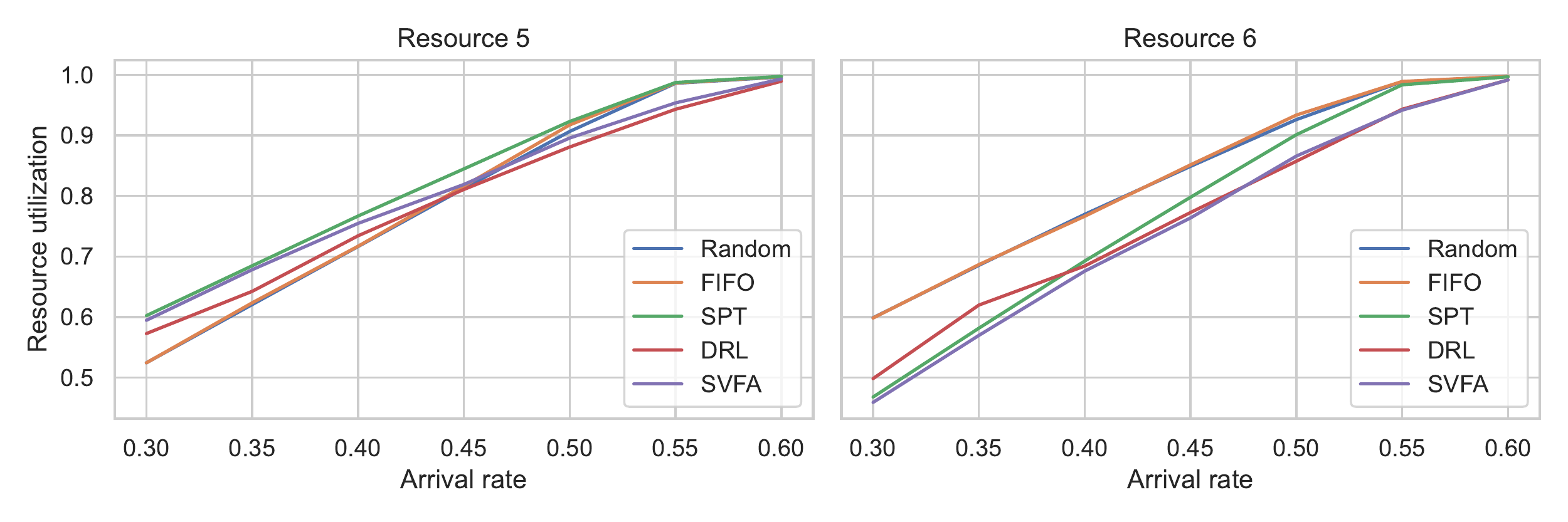}
    \caption{Utilization rate of the resources in the slow server scenario.}
    \label{fig:utilization_slow_server}
\end{figure}

As can be seen in Figure~\ref{fig:utilization_slow_server}, the resource utilization approaches 1 at $\lambda=0.55$, which shows that the process indeed becomes unstable at this arrival rate as the resources are always occupied. We found that using an arrival rate of $\lambda=0.5$ provides a challenging problem while keeping the system stable. Interestingly, Figure~\ref{fig:utilization_slow_server} and Figure~\ref{fig:varying_slow_server} show that a lower cycle time can be achieved with lower resource utilization, demonstrating the importance of an effective resource allocation policy in organizations. The behavior of the resources in other scenarios is the same as the one presented, where the utilization increases with the arrival rate in a similar trend.

To expand our experiments, we evaluated our models using an arrival pattern as described in Section~\ref{sec:evaluation protocol}. To account for the seasonality in arrivals, we trained our DRL model in an additional configuration, DRL$_\text{t}$, containing a feature representing the current progress in the arrival pattern. The results of our methods compared to three heuristics are shown in Table~\ref{tab:results scenarios pattern}.

\begin{table}[h]
    \centering
    \caption{Mean cycle time and CIs for the six scenarios using an arrival pattern.}
    \resizebox{1\linewidth}{!}{
    \begin{tabular}{lcccccc}
        \toprule
        Business process   & DRL & DRL$_\text{t}$ & SVFA & SPT & FIFO & Random \\
        \midrule
        Low utilization & 17.5 (0.37) & \textbf{17.2 (0.40)} & 17.9 (0.44) & \textbf{16.8 (0.42)} & 17.4 (0.4) & 21.4 (0.43) \\
        High utilization & \textbf{45.0 (1.58)} & 47.9 (3.07) & \textbf{44.5 (1.02)} & \textbf{44.4 (1.33)} & 50.6 (2.11) & 63.9 (3.73) \\
        Slow server & \textbf{35.2 (0.88)} & \textbf{34.6 (0.79)} & 39.2 (0.87) & 69.9 (2.87) & 41.6 (1.35) & 49.8 (2.26) \\
        Slow downstream & \textbf{28.8 (0.57)} & \textbf{29.1 (0.64)} & 37.5 (0.86) & 49.4 (1.15) & \textbf{28.4 (0.58)} & 33.6 (0.64) \\
        N-network & 18.4 (0.45) & 18.4 (0.38) & \textbf{13.2 (0.29)} & 18.6 (0.53) & \textbf{12.9 (0.29)} & 15.7 (0.37) \\
        Parallel & 35.4 (0.79) & \textbf{29.4 (0.58)} & 43.5 (1.01) & 48.2 (1.14) & \textbf{28.5 (0.67)} & 32.7 (0.68) \\
        \bottomrule
    \end{tabular}
    }
    \label{tab:results scenarios pattern}
\end{table}

The table shows that our DRL method performs similarly when using an arrival pattern compared to using a constant arrival rate. DRL and DRL$_\text{t}$ outperform or are competitive with the best-performing other methods in the first four scenarios. However, DRL$_\text{t}$ slightly outperforms DRL in the low utilization scenario and vice versa in the high utilization scenario. Interestingly, DRL$_\text{t}$ performs comparatively with FIFO in the parallel scenarios, whereas our DRL method was performing worse in the scenario with a constant arrival rate. These results show that under a dynamic arrival rate, our DRL method can still learn effective policies. In the scenarios, the additional temporal feature did not significantly improve the method's performance, except for the parallel scenario.

The SVFA method performed slightly worse under the arrival pattern, as it is only the (tied) best-performing method in two out of the six scenarios. However, in the other scenarios, it can still learn a policy that outperforms some of the heuristics. The results indicate that for both our methods, it is more challenging to learn a good policy when using dynamic arrivals compared to when using a constant arrival rate.

In the second part of the evaluation, we evaluated our methods on the composite processes using varying and dynamic arrival rates. Figure~\ref{fig:varying_complete_processes} shows the performance of our methods using varying arrival rates in the composite processes. Arrival rates above $\lambda=0.5$ are excluded to improve the figure's clarity and because the processes are unstable above this rate.

\begin{figure}[h]
    \centering
    \begin{subfigure}[b]{0.49\textwidth}
        \centering
        \includegraphics[width=\textwidth]{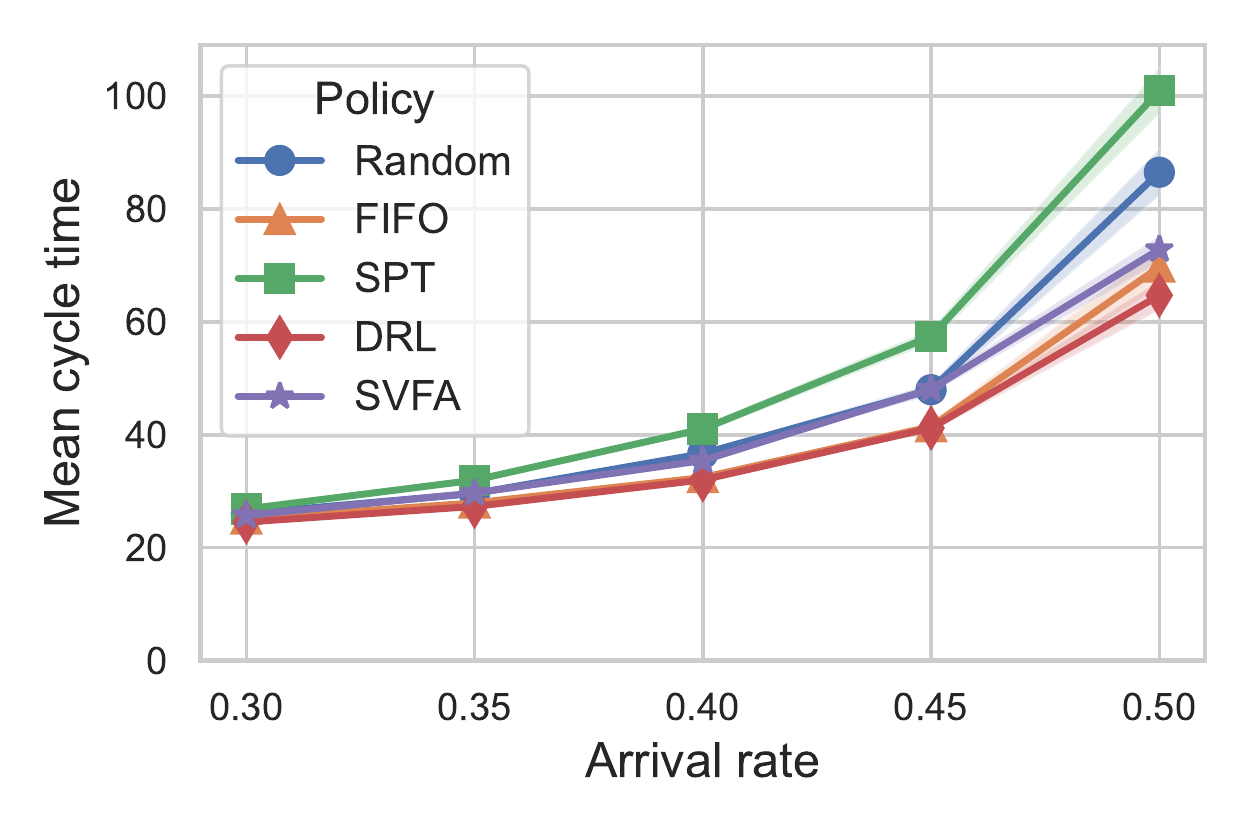}
        \caption{Composite}
        \label{fig:varying_complete}
    \end{subfigure}
    \hfill
    \begin{subfigure}[b]{0.49\textwidth}
        \centering
        \includegraphics[width=\textwidth]{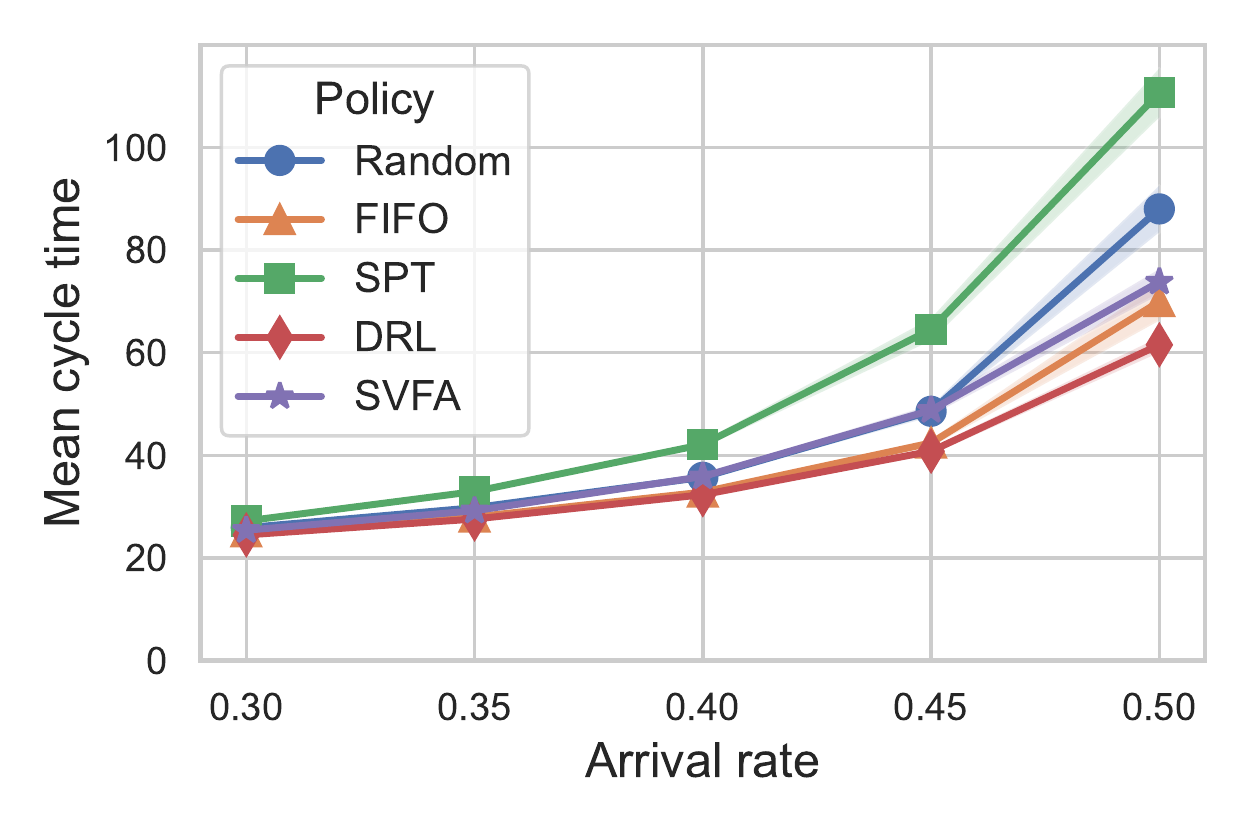}
        \caption{Composite reversed}
        \label{fig:varying_complete_reversed}
    \end{subfigure}
    \vfill
    \begin{subfigure}[b]{0.49\textwidth}
        \centering
        \includegraphics[width=\textwidth]{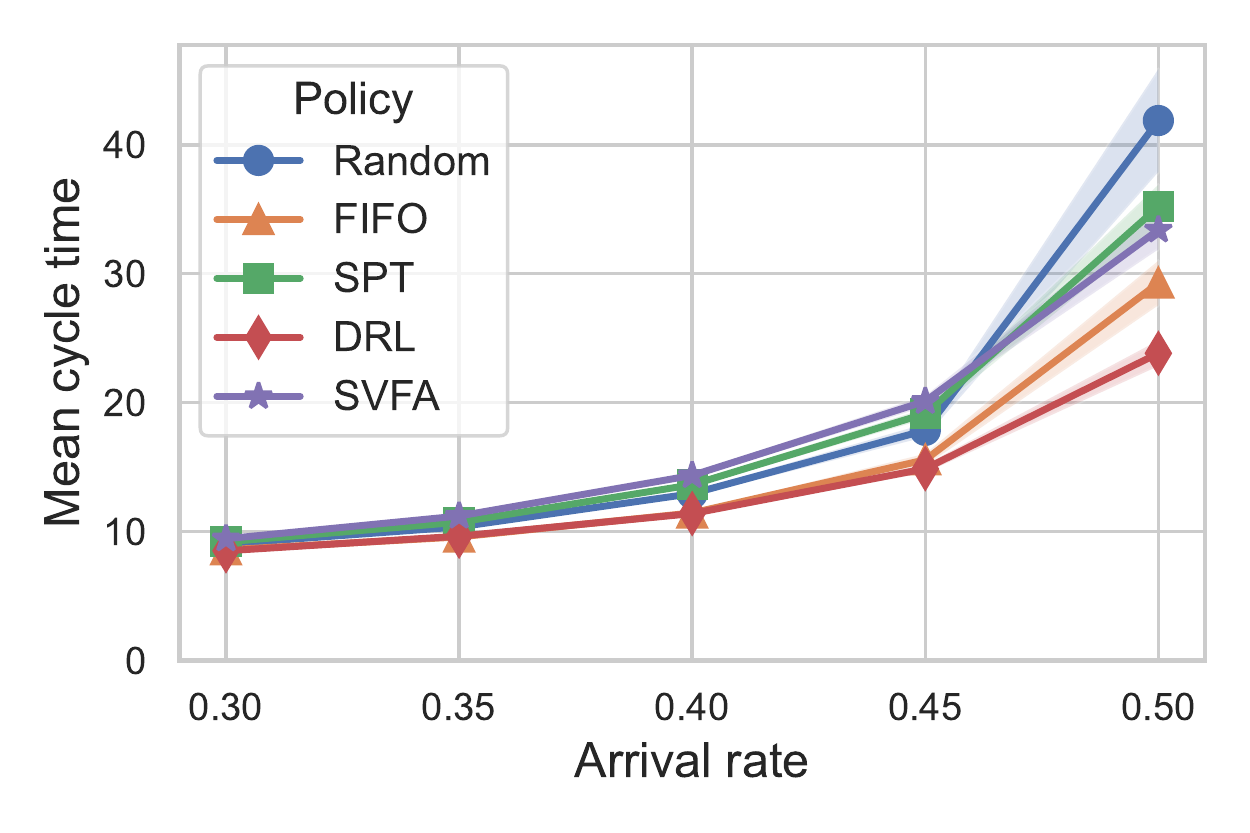}
        \caption{Composite parallel}
        \label{fig:varying_complete_parallel}
    \end{subfigure}
    \caption{Comparison of policies in the composite business processes using a range of arrival rates.}
    \label{fig:varying_complete_processes}
\end{figure}

The figure shows that the relative performance of each method remains consistent across different arrival rates. Our DRL method consistently outperforms or matches the performance of other methods across all rates. At lower arrival rates, DRL performs similarly to the FIFO policy. This is because, under low load conditions (i.e., low arrival rates), there is minimal queuing, making policies that prioritize case completions effective regardless of the resources executing the activity instances. Both FIFO and our DRL method adopt this policy. However, as the load increases ($\lambda=0.5$), the performance of FIFO and DRL diverges. The DRL method learns to balance prioritizing case completions with efficient resource utilization, a balance that the FIFO policy fails to achieve.

In both the composite and composite reversed processes, our SVFA method performs comparably to a random policy at arrival rates below $\lambda=0.5$, and slightly worse than the best-performing methods. The random policy’s performance at lower arrival rates suggests that strategic resource allocation policies offer limited benefits in such scenarios, making a random policy potentially sufficient. However, at $\lambda=0.5$, the performance of SVFA diverges from the random policy, indicating that correct resource allocation can significantly impact cycle time.

Finally, Table~\ref{tab:results complete pattern} presents the performance of each method and benchmark on composite processes using an arrival pattern. Both DRL and SVFA were outperformed by the FIFO heuristics. However, incorporating the feature that represents progress in the arrival pattern into DRL$\text{t}$ significantly enhanced the method's performance. DRL$_\text{t}$ outperformed all other methods in the composite parallel process and was competitive with FIFO in the composite process. Nonetheless, DRL$_\text{t}$ could not find a policy that matched FIFO in the composite reversed process.

\begin{table}[h]
\caption{Mean cycle time and CIs for the composite processes using an arrival pattern.}
\label{tab:results complete pattern}
\resizebox{1\linewidth}{!}{
\begin{threeparttable}
\begin{tabular}{lcccccc}
\toprule
Business process        & DRL       & DRL$_\text{t}$           & SVFA                  & SPT                  & FIFO                 & Random  \\ \midrule
        Composite & 120.9 (3.88) & \textbf{98.3 (3.37)} & 111.1 (2.99) & 128.9 (4.0) & \textbf{96.8 (2.95)} & 111.8 (3.39) \\
        Composite reversed & 126.4 (3.76) & 107.6 (3.66) & 121.1 (3.75) & 179.1 (5.73) & \textbf{96.3 (3.51)} & 113.0 (3.69) \\
        Composite parallel & 63.2 (1.73) & \textbf{51.5 (1.63)} & 85.4 (1.83) & 73.1 (1.85) & 55.1 (2.28) & 68.6 (3.19) \\
\bottomrule
\end{tabular}
\end{threeparttable}
}
\end{table}

Our methods, under a constant arrival rate, can learn policies that outperform or match the best benchmarks in archetypal business processes (Table~\ref{tab:results scenarios}). In all composite processes, our DRL method finds a policy with the lowest cycle time (Table~\ref{tab:results complete}).

When using arrival patterns, our methods perform similarly in most scenarios, though slightly worse in some, compared to a constant arrival rate (Table~\ref{tab:results scenarios pattern}). DRL$_\text{t}$ did not significantly improve DRL’s performance, except in the parallel scenario. In the composite parallel process, DRL$_\text{t}$ outperformed the best-performing benchmark and matched the best-performing benchmark in the composite model (Table~\ref{tab:results complete pattern}). In the composite reversed process, FIFO performed best. Our methods are designed for constant arrival rates, and while we added a simple feature to represent the arrival pattern improved performance, further research is needed to optimize policies for dynamic processes with, for example, seasonality in arrival rates or resource downtime.}

\section{Conclusion}\label{sec:conclusion}
\btext{This paper introduced two learning methods for resource allocation in business processes: deep reinforcement learning (DRL) and score-based value function approximation (SVFA). We evaluated these methods on six archetypal business processes, or scenarios, and three composite processes, which are a combination of these scenarios in sequence, sequence reversed, and parallel.

Under a constant arrival rate, our methods outperformed or matched the benchmarks in five out of six scenarios. In the scenario with parallelism, our methods could not compete with the FIFO heuristic.
\bbtext{The FIFO policy is a suitable and computationally efficient policy for resource allocation and its main strengths lie in prioritizing the quick completion of individual cases. This policy is especially effective in processes where choosing which activity to complete is more important than which resource to assign to it (i.e., slow downstream and parallelism scenario). Furthermore, our learning-based methods do not consider the prefix of a case, which could significantly improve the policy for the parallelism scenario. However, the disadvantage of FIFO, and heuristics in general, is that it can force sub-optimal assignments, while learning-based methods can adapt to different process states.} The scenario results showed that our learning-based methods are suitable for learning policies for process patterns commonly seen in business processes.

In the composite processes, our DRL method achieved a, on average, 12.7\% lower cycle time compared to the best benchmark. Our SVFA method could not find a competitive policy and performed slightly worse than the best-performing benchmark. While some heuristics exhibited good performance in some scenarios, they could not handle the complexity of all scenarios combined. On the other hand, DRL automatically learns important features from the current state of the process, independent of the process characteristics, making it adaptive to different business process patterns and flows. 

To investigate our methods further, we analyzed their performance when using an arrival pattern that adapts the arrival rate over time. The results showed that DRL and SVFA performed best in three and two scenarios, respectively. However, in the composite processes, our methods could not find a competitive policy. Adding a feature to DRL to represent the arrival pattern significantly improved the performance in the parallel scenario and the composite processes. After introducing this feature, DRL outperformed or matched the performance of the best-performing method in two out of three composite processes. 

In future work, the current methods can be improved and extended in several ways. First, in this paper, we only considered a single objective, but other or multiple objectives, such as case outcome, can be more fitting depending on the process. 
Second, as shown in the results, our proposed methods cannot sufficiently handle an arrival pattern. Our methods were designed for constant arrival rates, and while adding a simple feature representing the arrival pattern significantly reduced the cycle time, more research is needed to handle more complex dynamics, such as seasonality in the process. \bbtext{Furthermore, additional scenarios should be researched to further investigate the adaptability of our methods.
While we have already applied our DRL method to real-world business processes \cite{meneghello24}, we have not yet studied changes in the process over time, such as adding new employees.}
Third, process mining research has shown that the execution history of a case, or prefix, contains valuable information about future states. However, in current BPO methods, this information is not used and can be exploited to improve the optimization process. 
Lastly, in the N-network, the reward function encouraged the agent to focus on completing one of the two activities, resulting in high rewards, but also high cycle times. In future work, we aim to address the bias induced by the reward function and investigate a reward function that is more representative of the objective and does not require reward engineering.}

\hfill \break \noindent \textbf{Reproducibility.} The source code to reproduce the experiments can be found in our Github repository: \href{http://www.github.com/jeroenmiddelhuis/LearningResourceAllocation}{http://www.github.com/jeroenmiddelhuis/\\LearningResourceAllocation}.

\hfill \break \noindent \textbf{Acknowledgements.} The research leading up to this paper is supported by the Dutch foundation for scientific research (NWO) under the CERTIF-AI project (grant nr. 17998).



\bibliographystyle{elsarticle-num} 
\bibliography{references.bib}





\end{document}